\documentclass[11pt,a4paper]{scrreprt} 

\usepackage[utf8]{inputenc}
\usepackage[T1]{fontenc}
\usepackage{url}
\usepackage{booktabs}
\usepackage{amsfonts}
\usepackage{nicefrac}
\usepackage{microtype}
\usepackage{xcolor}
\usepackage{mathtools}
\usepackage{amsmath}
\usepackage{amsthm}
\usepackage{tcolorbox}
\usepackage{placeins}
\usepackage[shortlabels]{enumitem}

\usepackage[round]{natbib}

\usepackage[top=3cm, bottom=3cm, left=1.5cm, right=1.5cm]{geometry}
\usepackage{color}
\usepackage{amsmath}
\usepackage{amssymb}
\usepackage{amsfonts}
\usepackage{tcolorbox}
\usepackage[english]{babel}
\usepackage{fancyhdr}
\usepackage[mathscr]{euscript}

\usepackage{authblk,amssymb,latexsym,amsfonts,amsmath}
\usepackage{color}
\usepackage{algorithm}
\usepackage{algpseudocode}
\usepackage{bm}
\usepackage{mathtools}
\usepackage[title]{appendix}
\usepackage[T1]{fontenc}
\usepackage{dsfont}
\usepackage{natbib}
\usepackage{url}
\usepackage{float}
\usepackage{placeins}

\usepackage[font=small,labelfont=bf]{caption}
\usepackage{authblk} 

\fancypagestyle{firststyle}
{
   \fancyhf{}
   
}

\renewenvironment{proof}[1][\proofname]{\noindent{\bfseries #1.} }{\qed \\ }

\DeclareMathAlphabet{\mathcal}{OMS}{cmsy}{m}{n}

\usepackage[utf8]{inputenc}
\usepackage[T1]{fontenc}
\usepackage{url}
\usepackage{booktabs}
\usepackage{nicefrac}
\usepackage{microtype}
\usepackage{xcolor}
\usepackage{float}
\usepackage{amssymb,amsmath}
\usepackage{bm}
\usepackage{mathtools}
\usepackage[round]{natbib}
\usepackage{todonotes}
\setuptodonotes{inline}
\usepackage{verbatim}
\usepackage{placeins}
\usepackage{xspace}
\setlist[itemize]{leftmargin=2em}
\allowdisplaybreaks
\DeclareMathAlphabet\mathbfcal{OMS}{cmsy}{b}{n}

\newcommand{\coremmd}{h^{\operatorname{MMD}}_k}

\newcommand{\corehsick}{h^{\operatorname{HSIC}}_{\hsickx\!,\hsicky}}

\newcommand{\mmd}{\operatorname{MMD}_k}

\newcommand{\hsickx}{{k^{\mathcal{X}}}}
\newcommand{\hsicky}{{k^{\mathcal{Y}}}}

\newcommand{\hsicKx}{K^{\mathcal{X}}}
\newcommand{\hsicKy}{K^{\mathcal{Y}}}
\newcommand{\bhsicKx}{\bar{K}^{\mathcal{X}}}
\newcommand{\bhsicKy}{\bar{K}^{\mathcal{Y}}}

\newcommand{\one}{\bm{1}}

\newcommand{\Dcal}{{\mathcal{D}}}

\newcommand{\Xcal}{{\mathcal{X}}}
\newcommand{\Acal}{{\mathcal{A}}}
\newcommand{\Ycal}{{\mathcal{Y}}}
\newcommand{\Hcal}{{\mathcal{H}}}
\newcommand{\Hcald}{{\mathcal{H}^d}}
\newcommand{\Ccal}{{\mathcal{C}}}
\newcommand{\Kcal}{{\mathcal{K}}}

\newcommand{\Fcal}{{\mathcal{F}}}
\newcommand{\Gcal}{{\mathcal{G}}}
\newcommand{\Lcal}{{\mathcal{L}}}

\newcommand{\fbm}{{\bm{f}}}
\newcommand{\kbm}{{\bm{k}}}
\newcommand{\gbm}{{\bm{g}}}
\newcommand{\sbm}{{\bm{s}}}
\newcommand{\sbmp}{{\bm{s}_P}}
\newcommand{\sbmq}{{\bm{s}_Q}}
\newcommand{\dbm}{{\bm{\delta}}}
\newcommand{\xibm}{{\bm{\xi}}}

\newcommand{\E}{\mathbb{E}}
\newcommand{\R}{\mathbb{R}}
\newcommand{\N}{\mathbb{N}}

\newcommand{\Eb}[1]{\mathbb{E}\!\left[#1\right]}

\newcommand{\pp}[1]{\!\left(#1\right)}
\newcommand{\cb}[1]{\!\left\{#1\right\}}

\newcommand{\bigO}[1]{\mathcal{O}\!\left(#1\right)}

\newcommand{\ie}{\emph{i.e.}\xspace}

\newcommand{\tr}{\mathrm{tr}}

\newcommand{\iid}{\overset{\mathrm{i.i.d.}}{\sim}}
\newcommand{\dd}{\mathrm{d}}

\DeclarePairedDelimiter\floor{\lfloor}{\rfloor}

\usepackage{mathtools}
\usepackage{xcolor}
\usepackage{lastpage}
\usepackage{enumitem}
\usepackage{bbm}
\usepackage{bm}
\usepackage{float}
\usepackage{caption}
\usepackage{setspace}
\usepackage{placeins}
\usepackage{afterpage}
\usepackage{multirow}
\usepackage{array}
\usepackage{tabularx}
\usepackage{hhline}
\usepackage{colortbl} 
\usepackage{booktabs}

\newcommand{\p}[1]{\!\left( #1 \right)}

\DeclareMathAlphabet{\mymathbb}{U}{BOONDOX-ds}{m}{n}

\DeclarePairedDelimiter\abs{\lvert}{\rvert}
\DeclarePairedDelimiter\norm{\lVert}{\rVert}
\makeatletter
\let\oldabs\abs
\def\abs{\@ifstar{\oldabs}{\oldabs*}}
\let\oldnorm\norm
\def\norm{\@ifstar{\oldnorm}{\oldnorm*}}
\makeatother

\makeatletter
\newcommand*\bigcdot{\mathpalette\bigcdot@{.5}}
\newcommand*\bigcdot@[2]{\mathbin{\vcenter{\hbox{\scalebox{#2}{$\m@th#1\bullet$}}}}}
\makeatother

\makeatletter
\newcommand\footnoteref[1]{\protected@xdef\@thefnmark{\ref{#1}}\@footnotemark}
\makeatother

\algblockdefx{MRepeat}{EndRepeat}{\textbf{repeat}}{}
\algnotext{EndRepeat}

\makeatletter
\newcommand\fs@nobottomruled{\def\@fs@cfont{\bfseries}\let\@fs@capt\floatc@ruled
  \def\@fs@pre{\hrule height.8pt depth0pt \kern2pt}\def\@fs@post{}\def\@fs@mid{\kern2pt\hrule\kern2pt}\let\@fs@iftopcapt\iftrue}
\makeatother

\floatstyle{nobottomruled}
\restylefloat{algorithm}

\makeatletter
\AtBeginDocument{\def\HH@loop{\ifx\@tempb`\def\next##1{\the\toks@\cr}\else\let\next\HH@let
  \ifx\@tempb|\if@tempswa
          \ifx\CT@drsc@\relax
           \HH@add{\hskip\doublerulesep}\else
           \HH@add{{\CT@drsc@\vrule\@width\doublerulesep}}\fi
          \fi\@tempswatrue
          \HH@add{{\CT@arc@\vline}}\else
  \ifx\@tempb:\if@tempswa
          \ifx\CT@drsc@\relax
           \HH@add{\hskip\doublerulesep}\else
           \HH@add{{\CT@drsc@\vrule\@width\doublerulesep}}\fi
              \fi\@tempswatrue
      \HH@add{\@tempc\HH@box\arrayrulewidth\arrayrulewidth\@tempc}\else
  \ifx\@tempb;\if@tempswa
          \ifx\CT@drsc@\relax
           \HH@add{\hskip\doublerulesep}\else
           \HH@add{{\CT@drsc@\vrule\@width\doublerulesep}}\fi
              \fi\@tempswatrue
      \HH@add{\@tempc\HH@box\z@\arrayrulewidth\@tempc}\else
  \ifx\@tempb##\if@tempswa\HH@add{\hskip\doublerulesep}\fi\@tempswatrue
         \HH@add{{\CT@arc@\vline\copy\@ne\@tempc\vline}}\else
  \ifx\@tempb~\@tempswafalse
           \if@firstamp\@firstampfalse\else\HH@add{&\omit}\fi
              \ifx\CT@drsc@\relax
                \HH@add{\hfil}\else
                 \HH@add{{\CT@drsc@\leaders\hrule\@height\HH@height\hfil}}\fi
                 \else
  \ifx\@tempb-\@tempswafalse
           \gdef\HH@height{\arrayrulewidth}\if@firstamp\@firstampfalse\else\HH@add{&\omit}\fi
              \HH@add{{\CT@arc@\leaders\hrule\@height\arrayrulewidth\hfil}}\else
  \ifx\@tempb=\@tempswafalse
       \gdef\HH@height{\dimen\thr@@}\if@firstamp\@firstampfalse\else\HH@add{&\omit}\fi
       \HH@add
          {\rlap{\copy\@ne}\leaders\copy\@ne\hfil\llap{\copy\@ne}}\else
  \ifx\@tempb t\HH@add{\def\HH@height{\dimen\thr@@}\HH@box\doublerulesep\z@}\@tempswafalse\else
  \ifx\@tempb b\HH@add{\def\HH@height{\dimen\thr@@}\HH@box\z@\doublerulesep}\@tempswafalse\else
  \ifx\@tempb>\def\next##1##2{\HH@add{{\baselineskip\p@\relax
       ##2\global\setbox\@ne\HH@box\doublerulesep\doublerulesep}}\HH@let!}\else
  \PackageWarning{hhline}{\meaning\@tempb\space ignored in \noexpand\hhline argument\MessageBreak}\fi\fi\fi\fi\fi\fi\fi\fi\fi\fi\fi
  \next}
}

\usepackage[utf8]{inputenc}
\usepackage[T1]{fontenc}
\usepackage{url}
\usepackage{booktabs}
\usepackage{amsfonts}
\usepackage{nicefrac}
\usepackage{microtype}
\usepackage{xcolor}
\usepackage{subcaption}
\usepackage{microtype}
\usepackage{graphicx}
\usepackage{booktabs} 
\usepackage{amsmath}
\usepackage{amssymb}
\usepackage{bm}
\usepackage{mathtools}
\usepackage{xspace}
\usepackage{bbm}

\newcommand{\kl}{k_{\lambda}}

\newcommand{\km}{k_{\lambda}^\Xcal}
\newcommand{\kn}{k_{\mu}^\Ycal}

\usepackage[utf8]{inputenc}
\usepackage[T1]{fontenc}
\usepackage{url}
\usepackage{booktabs}
\usepackage{amsfonts}
\usepackage{nicefrac}
\usepackage{microtype}
\usepackage{xcolor}
\usepackage{mathtools}
\usepackage{amsmath}
\usepackage{tcolorbox}
\usepackage{placeins}
\usepackage{amssymb}
\usepackage{bm}
\usepackage{float}

\newcommand{\D}{\mathcal{D}}

\newcommand{\abss}[1]{\big|#1\big|}

\usepackage[utf8]{inputenc} 
\usepackage[T1]{fontenc}    
\usepackage{url}            
\usepackage{booktabs}       
\usepackage{amsfonts}       
\usepackage{nicefrac}       
\usepackage{microtype}      
\usepackage{xcolor}         

\usepackage{amssymb,amsmath}
\usepackage{bm}
\usepackage{mathtools}
\usepackage{verbatim}
\usepackage{enumitem}
\usepackage{placeins}

\usepackage{wrapfig}

\usepackage{thm-restate}

\RequirePackage[
  pdfstartview=FitH,
  breaklinks=true,
  bookmarks=true,
  colorlinks=true,
  linkcolor= blue,
  anchorcolor=blue,
  citecolor=blue,
  filecolor=blue,
  menucolor=blue,
  urlcolor=blue
  ]{hyperref}
  \AtBeginDocument{
  \hypersetup{
    pdfauthor = {Antonin Schrab},
    colorlinks = true,
    urlcolor = cyan,
    linkcolor = blue,
    citecolor = teal,
    pdftitle = {Title - compilation : \today}
  }
}
\usepackage{hyperref} 
\usepackage{cleveref}
\crefname{assumption}{Assumption}{Assumptions}
\crefname{equation}{Equation}{Equations}
\crefname{figure}{Figure}{Figures}
\crefname{table}{Table}{Tables}
\crefname{section}{Section}{Sections}
\crefname{theorem}{Theorem}{Theorems}
\crefname{lemma}{Lemma}{Lemmas}
\crefname{corollary}{Corollary}{Corollaries}
\crefname{example}{Example}{Examples}
\crefname{appendix}{Appendix}{Appendices}
\crefname{remark}{Remark}{Remarks}

\usepackage[nottoc]{tocbibind}

\newcommand{\mysection}[1]{\section{#1}}
\newcommand{\mysubsection}[1]{\subsection{#1}}

\makeatletter 
 
\@addtoreset{algorithm}{chapter} 
\makeatother

\creflabelformat{equation}{#2\textup{#1}#3}
\crefformat{footnote}{#2\footnotemark[#1]#3}

\usepackage{thmtools}
\declaretheorem[name=Theorem,numberwithin=chapter]{theorem}
\declaretheorem[name=Proposition,sibling=theorem]{proposition}
\declaretheorem[name=Corollary,sibling=theorem]{corollary}

\declaretheorem[name=Lemma,sibling=theorem]{lemma}

\counterwithout*{footnote}{chapter}

\counterwithout{equation}{chapter} 

\begin{document}

\thispagestyle{firststyle}
{\center

	{
		\fontsize{32}{52}\selectfont 
		\vspace{2cm}
		\textbf{A Practical Introduction to}\\[-0.5cm]
		\textbf{Kernel Discrepancies: }\\[-0.5cm]
		\textbf{MMD, HSIC \& KSD}\\[1cm]
	}

	{
		\fontsize{16}{14}\selectfont 
		\textbf{Antonin Schrab} \\[0.3cm]
		\textit{Centre for Artificial Intelligence } \\[0.2cm]
		\textit{Gatsby Computational Neuroscience Unit} \\[0.2cm]
		\textit{University College London} \\[0.3cm]
	}

}

\vskip 0.5cm
{\LARGE\centerline{\textbf{Abstract}}}
\vskip 0.3cm

\begin{center}
\begin{minipage}{8.0cm}
This article provides a practical introduction to kernel discrepancies, focusing on the Maximum Mean Discrepancy (MMD), the Hilbert--Schmidt Independence Criterion (HSIC), and the Kernel Stein Discrepancy (KSD). Various estimators for these discrepancies are presented, including the commonly-used V-statistics and U-statistics, as well as several forms of the more computationally-efficient incomplete U-statistics. The importance of the choice of kernel bandwidth is stressed, showing how it affects the behaviour of the discrepancy estimation. Adaptive estimators are introduced, which combine multiple estimators with various kernels, addressing the problem of kernel selection.
\end{minipage}
\end{center}
\bigskip
\bigskip

This paper corresponds to the introduction of my PhD thesis \citep[Chapter 2]{schrab2025optimal} and is presented as a standalone article to introduce the reader to kernel discrepancies estimators.
First, in \Cref{sec:kernels}, we define kernels, Reproducing Kernel Hilbert Spaces, mean embeddings and cross-covariance operators, and present kernel properties such as characteristicity, universality and translation invariance.
Then, in \Cref{sec:kernel_discrepancies}, we introduce the Maximum Mean Discprecancy, the Hilbert--Schmidt Independence Criterion, and the Kernel Stein Discrepancy, as well as their estimators, and we discuss the importance of the choice of kernel for such measures.
We then introduce a collection of statistics in
\Cref{sec:fixed_estimators}, including the commonly-used complete statistics, as well as their incomplete counterparts which trade accuracy for computational efficiency.
Finally, in \Cref{sec:adaptive_estimators}, we construct adaptive estimators combining multiple statistics with various kernels, which is one method to address the problem of kernel selection.

\mysection{Kernels}
\label{sec:kernels}

In this section, 
we introduce kernels and RKHSs, 
define characteristicity and universality of kernels,
present translation-invariant kernels with bandwidth parameters,
and provide some classical kernel examples.

We refer the readers 
to \citet{sejdinovic2012rkhs,gretton2013introduction} for some kernel/RKHS introductions,
to \citet{azonszajn1950theory,steinwart2008support,berlinet2011reproducing} for details about the RKHS constructions,
to \citet{fukumizu2004dimensionality,sriperumbudur2008injective,sriperumbudur2010relation,sriperumbudur2010hilbert,sriperumbudur2011universality,carmeli2010vector} for the characteristicity and universality of kernels,
and to \citet{muandet2017kernel} for an in-depth review of kernel mean embedding methods.

\paragraph{Kernel \& RKHS.}
We present three definitions of a \emph{`kernel'} and then discuss their relations.
First, recall that a vector space $\Hcal$ is called a Hilbert space if it is equipped with an inner product $\langle\cdot,\cdot\rangle_{\Hcal}$ and is complete (\ie every Cauchy sequence converges with respect to the metric induced by the inner product). 
In the following three definitions, let $\Xcal$ to be a non-empty set. 

\begin{enumerate}
\item A function $k\colon \Xcal\times\Xcal \to \R$ is a \emph{kernel} if there exist a Hilbert space $\Hcal$ and a function $\phi\colon \Xcal\to\Hcal$ (called \emph{feature map}) such that
$$
k(x,y) = \langle \phi(x), \phi(y)\rangle_{\Hcal}
$$
for all $x,y\in\Xcal$.
\item A symmetric function $k\colon \Xcal\times\Xcal \to \R$ is a \emph{positive definite kernel} if
$$
\sum_{i=1}^n \sum_{j=1}^n c_i c_j k(x_i,x_j) \geq 0
$$
for all $x_1,\dots,x_n\in\Xcal$ and all $c_1,\dots,c_n\in\R$, for any $n\in\N$.
\item For a Hilbert space $\Hcal$ of real-valued functions on $\Xcal$, a function $k\colon \Xcal\times\Xcal \to \R$ is a \emph{reproducing kernel} of $\Hcal$ if 
\begin{itemize}
	\item $k(x,\cdot)\in\Hcal$ for all $x\in\Xcal$,
	\item $\langle f, k(x,\cdot)\rangle_{\Hcal} = f(x)$ for all $f\in\Hcal$ and all $x\in\Xcal$.
\end{itemize}
The resulting space $\Hcal$ is then called a \emph{Reproducing Kernel Hilbert Space} (RKHS).
\end{enumerate}

\noindent
These three definitions are equivalent in the following sense.

\begin{itemize}
	\item $[1\Longrightarrow2]$ Every kernel is a positive definite kernel as $\sum_{i=1}^n \sum_{j=1}^n c_i c_j k(x_i,x_j) = \|\sum_{i=1}^nc_i\phi(x_i)\|_{\Hcal}\geq 0$.
	\item $[2\Longrightarrow1]$ Every positive definite kernel is guaranteed to be an inner product between features in an Hilbert space \citep[Theorem 4.16]{steinwart2008support}.
	\item $[2\Longrightarrow3]$ Given a positive definite kernel $k$ on $\Xcal\times\Xcal$, there exists a unique Hilbert space $\Hcal_k$ of real-valued functions on $\Xcal$ for which $k$ is the reproducing kernel, $\Hcal_k$ is the (unique) RKHS associated to $k$ \citep[Moore-Aronsajn Theorem]{azonszajn1950theory}.
	\item $[3\Longrightarrow1]$ A reproducing kernel is a kernel with feature map $\phi(x)=k(x,\cdot)$ as $\langle k(x,\cdot), k(y,\cdot)\rangle_{\Hcal} = k(x,y)$.
\end{itemize}

\noindent
There exists an equivalent definition of an RKHS, which perhaps suprisingly, does not involve the notion of a kernel. 

\begin{enumerate}
	\item[4.] A Hilbert space $\Hcal$ of real-valued functions on $\Xcal$ is an RKHS if the evaluation functional is continuous, that is, $|f(x)| \leq C_x\|f\|_{\Hcal}$ for some $C_x>0$, for all $f\in\Hcal$ and all $x\in\Xcal$.
\end{enumerate}

\noindent
These two RKHS definitions are indeed equivalent.

\begin{itemize}
\item $[3\Longrightarrow4]$ Using the reproducing property followed by Cauchy--Schwartz inequality, we obtain $|f(x)| \leq \sqrt{k(x,x)}\|f\|_{\Hcal}$ for all $f\in\Hcal$ and all $x\in\Xcal$.
\item $[4\Longrightarrow3]$ By Riesz Representation Theorem \citep[Theorem 4.12]{rudin1987real}, for all $x\in\Xcal$ there exists $g_x\in\Hcal$ such that $f(x) = \langle f,g_x\rangle_{\Hcal}$ for all $f\in\Hcal$. The function $k(x,y) \coloneqq g_x(y)$ satisfies the two properties of definition 3 and hence is a repoducing kernel.
\end{itemize}

\noindent
We note that, from the reproducing property (definition 3 above), it also follows that (\citealp[Theorem 1,][]{zhou2008derivative} and \citealp[Lemma C.9,][]{barp2022targeted})
\begin{equation}
\label{eq:reproducing_derivative}
\textstyle\frac{\partial}{\partial x}f(x) 
= 
\big\langle f, \frac{\partial}{\partial x}k(x,\cdot)\big\rangle_{\Hcal}
\end{equation}
for all $f\in\Hcal$ and all $x\in\Xcal$, under appropriate regularity conditions.

\paragraph{Kernel mean embedding.}
Let $P$ be a probability distribution on $\Xcal$.
Riesz Representation Theorem \citep[Theorem 4.12]{rudin1987real} guarantees the existence of a unique element $\mu_P\in\Hcal$ satisfying
$$\langle f, \mu_P\rangle_{\Hcal}= \E_P[f(X)]= \E_P\!\left[\langle f, k(X,\cdot)\rangle_{\Hcal}\right]
$$
for all $f\in\Hcal$, where $\E_P[f(X)]$ denotes the expectation of $f$ with respect to $P$.
This element $\mu_P$ is called the \emph{kernel mean embedding} and can be written as
$$
\mu_P = \E_P[k(X,\cdot)]
$$
so that
$
\langle f, \E_P[k(X,\cdot)]\rangle_{\Hcal}
= \E_P\!\left[\langle f, k(X,\cdot)\rangle_{\Hcal}\right]
$
is justisfied.
Abusing notation, we write $\frac{\partial}{\partial X}f(X)$ for $\frac{\partial}{\partial x}f(x)\big|_{x=X}$,
we then have
$$
\E_P\!\left[\textstyle\frac{\partial}{\partial X}f(X)\right]
=
\E_P\!\left[\langle f, \textstyle\frac{\partial}{\partial X}k(X,\cdot)\rangle_{\Hcal}\right]
= 
\left\langle f, \E_P\!\left[\textstyle\frac{\partial}{\partial X}k(X,\cdot)\right]\right\rangle_{\Hcal}
$$
where the existence of $\E_P\!\left[\textstyle\frac{\partial}{\partial X}k(X,\cdot)\right]$ is again guaranteed by Riesz Representation Theorem.

\paragraph{Cross-covariance operator.}
Given two Hilbert spaces $\Hcal_1$ and $\Hcal_2$ with associated inner products $\langle\cdot,\cdot\rangle_{\Hcal_1}$ and $\langle\cdot,\cdot\rangle_{\Hcal_2}$, respectively, the tensor product $\otimes$ and Hilbert--Schmidt inner product $\langle\cdot,\cdot\rangle_{\mathrm{HS}}$ are defined as $\langle f\otimes g, \tilde{f}\otimes \tilde{g}\rangle_{\mathrm{HS}} = \langle f,\tilde{f}\rangle_{\Hcal_1}\langle g,\tilde{g}\rangle_{\Hcal_2}$ for all $f,\tilde{f}\in\Hcal_1$ and all $g,\tilde{g}\in\Hcal_2$.
Let $P_{XY}$ be a joint probability distribution on $\Xcal\times\Ycal$, and consider kernels $\hsickx$ and $\hsicky$ on $\Xcal\times\Xcal$ and $\Ycal\times\Ycal$, respectively.
The cross-covariance operator is defined as the linear operator $\Ccal_{P_{XY}}\colon\Hcal_{\hsicky}\to\Hcal_{\hsickx}$ satisfying
\begin{align*}
\langle f, \Ccal_{P_{XY}}g\rangle_{\Hcal_\hsickx}
	&=
\E_{P_{XY}}\!\big[
	\left( f(X) - \E_{P_X}[f(X')] \right)
	\left( g(Y) - \E_{P_Y}[g(Y')] \right)
\big] \\
&= 
\E_{P_{XY}}\!\big[
	\langle f, \hsickx(X,\cdot) - \mu_{P_X}\rangle_{\Hcal_\hsickx}
	\langle g, \hsicky(Y,\cdot) - \mu_{P_Y}\rangle_{\Hcal_\hsicky}
\big] \\
&= 
\E_{P_{XY}}\!\big[
	\left\langle 
	f \otimes g, 
	\big(\hsickx(X,\cdot) - \mu_{P_X}\big) \otimes \big(\hsicky(Y,\cdot) - \mu_{P_Y}\big)
	\right\rangle_{\mathrm{HS}}
\big] \\
&= 
\E_{P_{XY}}\!\left[
	\left\langle 
	f, 
	\Big(\big(\hsickx(X,\cdot) - \mu_{P_X}\big) \otimes \big(\hsicky(Y,\cdot) - \mu_{P_Y}\big)\Big)g
	\right\rangle_{\Hcal_\hsickx}
\right]
\end{align*}
for all $f\in\Hcal_k$ and all $g\in\Hcal_\hsicky$, where $\langle\cdot,\cdot\rangle_{\mathrm{HS}}$ denotes the Hilbert--Schmidt inner product.
The existence and uniqueness of $\Ccal_{P_{XY}}$ is guaranteed by Riesz Representation Theorem.
The notation 
$$
\Ccal_{P_{XY}} = \E_{P_{XY}}\!\left[ 
	\big(\hsickx(X,\cdot) - \mu_{P_X}\big)
	\otimes 
	\big(\hsicky(Y,\cdot) - \mu_{P_Y}\big)
\right]
$$
is justified in the sense that
\begin{align*}
&\left\langle f, \E_{P_{XY}}\!\left[ 
	\big(\hsickx(X,\cdot) - \mu_{P_X}\big)
	\otimes 
	\big(\hsicky(Y,\cdot) - \mu_{P_Y}\big)
\right]g\right\rangle_{\Hcal_\hsickx} \\
=\ 
&\E_{P_{XY}}\!\left[
	\left\langle 
	f, 
	\Big(\big(\hsickx(X,\cdot) - \mu_{P_X}\big) \otimes \big(\hsicky(Y,\cdot) - \mu_{P_Y}\big)\Big)g
	\right\rangle_{\Hcal_\hsickx}
\right]
\end{align*}
which gives
\begin{equation}
\label{eq:Cxy_property}
\begin{aligned}
&\left\langle f\otimes g, \E_{P_{XY}}\!\left[ 
	\big(\hsickx(X,\cdot) - \mu_{P_X}\big)
	\otimes 
	\big(\hsicky(Y,\cdot) - \mu_{P_Y}\big)
\right]\right\rangle_{\mathrm{HS}}\\
=\ 
&\E_{P_{XY}}\!\left[
	\left\langle 
	f \otimes g, 
	\big(\hsickx(X,\cdot) - \mu_{P_X}\big) \otimes \big(\hsicky(Y,\cdot) - \mu_{P_Y}\big)
	\right\rangle_{\mathrm{HS}}
\right].
\end{aligned}
\end{equation}

\paragraph{Characteristic kernel.}
A kernel $k$ is \emph{characteristic} \citep{fukumizu2004dimensionality,sriperumbudur2008injective,sriperumbudur2010hilbert,sriperumbudur2011universality} if $\mu_P = \mu_Q$ implies $P=Q$, where $\mu_P=\E_P[k(X,\cdot)]$. 
In other words, the kernel mean embedding captures all the information about the distribution, in the sense that, if two kernel mean embeddings are the same (\ie $\mu_P=\mu_Q$), then the distributions must be the same (\ie $P=Q$).

\paragraph{Universal kernel.}
A kernel $k\colon\Xcal\times\Xcal\to\R$ is $C_0$\emph{-universal} \citep{carmeli2010vector,sriperumbudur2010relation} if its associated RKHS $\Hcal_k$ is dense in $C_0(\Xcal,\R)$ (\ie the space of continuous functions from $\Xcal$ to $\R$ vanishing at infinity).
There exist various notions of universality, we refer the reader to \citet{sriperumbudur2011universality} for details.

\paragraph{Translation invariance, radial kernels \& bandwidths.}
A kernel $k\colon \R^d\times\R^d\to\R$ is \emph{translation-invariant} if 
$$
k(x,y) = \psi(x-y)
$$
for some (positive definite) function $\psi\colon\R^d\to\R$, often required to satisfy $\psi(0)=1$.
Then, for any \emph{bandwidth} $\lambda>0$, the scaled function
$$
k_\lambda(x,y) = \psi\pp{\frac{x-y}{\lambda}}
$$
is also a kernel as it is equal to $ k\pp{\frac{x}{\lambda},\frac{y}{\lambda}}$.
We say that $k\colon \R^d\times\R^d\to\R$ is a \emph{radial kernel}\footnote{In the literature, a radial kernel is sometimes defined only for the special case $r=2$.} if 
$$
k(x,y) = \Psi(\|x-y\|_r)
$$
for some $r\geq 1$ and some function $\Psi\colon\R\to\R$ with $\Psi(0)=1$, giving
$$
k_\lambda(x,y) = \Psi\pp{\frac{\|x-y\|_r}{\lambda}}.
$$
Note that $k_\lambda(x,x) = 1$ for all $x\in\R^d$ and all $\lambda>0$.
For $x\neq y$ both in $\R^d$, we have
\begin{equation}
	\label{eq:kernel_limit}
	k_\lambda(x,y) \to 0 \textrm{ as } \lambda\to0 
	\qquad\textrm{ and }\qquad
	k_\lambda(x,y) \to 1 \textrm{ as } \lambda\to\infty.
\end{equation}

\paragraph{Kernel examples.}
We now present some commonly used kernels which are characteristic and $C_0$-universal:
\newline
the \emph{Gausssian kernel}
$$
k_\lambda(x,y) = \exp\pp{-\frac{\|x-y\|_2^2}{\lambda^2}},
$$
the \emph{Laplace kernel}
$$
k_\lambda(x,y) = \exp\pp{-\frac{\|x-y\|_1}{\lambda}},
$$
the \emph{inverse multiquadric IMQ kernel}
$$
k_\lambda(x,y) 
= \frac{1}{\sqrt{\lambda^2+\|x-y\|_2^2}}
\propto \frac{1}{\sqrt{1+\displaystyle\frac{\|x-y\|_2^2}{\lambda^2}}},
$$
and the \emph{Mat\'ern kernels} with $\nu=0.5,1.5,2.5,3.5,4.5$ and $L^r$-distances for $r\geq 1$
$$
k_\lambda(x,y) = \exp\pp{-\frac{\|x-y\|_r}{\lambda}},
$$

$$
k_\lambda(x,y) = \pp{1+\sqrt{3}\frac{\|x-y\|_r}{\lambda}}\exp\pp{-\sqrt{3}\frac{\|x-y\|_r}{\lambda}},
$$

$$
k_\lambda(x,y) = \pp{1+\sqrt{5}\frac{\|x-y\|_r}{\lambda}+\frac{5}{3}\frac{\|x-y\|_r^2}{\lambda^2}}\exp\pp{-\sqrt{5}\frac{\|x-y\|_r}{\lambda}},
$$

$$
k_\lambda(x,y) = \pp{
	1+\sqrt{7}\frac{\|x-y\|_r}{\lambda}
	+\frac{2\cdot7}{5}\frac{\|x-y\|_r^2}{\lambda^2}
	+\frac{7\sqrt{7}}{3\cdot5}\frac{\|x-y\|_r^3}{\lambda^3}
}\exp\pp{
	-\sqrt{7}\frac{\|x-y\|_r}{\lambda}
},
$$

$$
k_\lambda(x,y) = \pp{
	1+3\frac{\|x-y\|_r}{\lambda}
	+\frac{3\cdot6^2}{28}\frac{\|x-y\|_r^2}{\lambda^2}
	+\frac{6^3}{84}\frac{\|x-y\|_r^3}{\lambda^3}
	+\frac{6^4}{1680}\frac{\|x-y\|_r^4}{\lambda^4}
}\exp\pp{
	-3\frac{\|x-y\|_r}{\lambda}
}.
$$

\mysection{Kernel discrepancies}
\label{sec:kernel_discrepancies}

We introduce the Maximum Mean Discrepancy (MMD) in \Cref{subsec:mmd}, the Hilbert--Schmidt Independence Criterion (HSIC) in \Cref{subsec:hsic}, and the Kernel Stein Discrepancy (KSD) in \Cref{subsec:ksd}.

\mysubsection{MMD: Maximum Mean Discrepancy}
\label{subsec:mmd}

\paragraph{MMD measure.}
As a measure between two probability distributions $P$ and $Q$, we consider the kernel-based \emph{Maximum Mean Discrepancy} (MMD---\citealp{GreBorRasSchetal07c,gretton2012kernel}). For a given RKHS $\mathcal{H}_k$ with reproducing kernel $k$, the MMD is defined as the integral probability metric \citep{muller1997integral}
\begin{equation}
\label{mmdagg_eq:mmd}
	\mathrm{MMD}_k(P, Q) 
	\coloneqq \sup_{f\in\mathcal{H}_k \,:\, \|f\|_{\mathcal{H}_k} \leq 1} 
	\mathbb{E}_{X\sim P}[f(X)] - \mathbb{E}_{Y\sim Q}  [f(Y)].
\end{equation}
We often simply write $\mathrm{MMD}_k$ for $\mathrm{MMD}_k(P, Q)$ when the distributions are clear from the context, and similarly for other discrepancies.
Using the reproducibility property, we obtain
\begin{equation}
	\mathrm{MMD}_k
	= \sup_{f\in\mathcal{H}_k \,:\, \|f\|_{\mathcal{H}_k} \leq 1} 
	\langle f,\mu_P-\mu_Q\rangle_{\Hcal_k}
	= \|\mu_P-\mu_Q\|_{\Hcal_k}
\end{equation}
and
\begin{equation}
\label{mmdagg_eq:mmd2}
	\mathrm{MMD}_k^2
	= \|\mu_P-\mu_Q\|_{\Hcal_k}^2
	= \E_{P,P}[k(X,X')] - 2 \E_{P,Q}[k(X,Y)] + \E_{Q,Q}[k(Y,Y')]
\end{equation}
by the properties of kernel mean embeddings, where $X,X'\sim P$ and $Y,Y'\sim Q$ are independent copies.
We observe that the MMD is the $\Hcal_k$-norm of the difference between the mean embeddings.
We note that the MMD can be leveraged to construct divergences for more general two-sample problems \citep{chau2024credal}.

\paragraph{MMD V-statistic.}
We now introduce some estimators of the MMD given some independent samples $X_1,\dots,X_m\iid P$ and $Y_1,\dots,Y_n\iid Q$.
We let $\widehat P$ and $\widehat Q$ denote the empirical distributions (uniform distributions on the datapoints).
The plug-in estimator \citep[Equations 2 and 5]{gretton2012kernel} for $\mathrm{MMD}^2_k(P,Q)$ is $\mathrm{MMD}^2_k(\widehat P,\widehat Q)$ which from \Cref{mmdagg_eq:mmd2} is equal to
\begin{equation}
	\label{mmdagg_mmdmnV}
	V_{\mathrm{MMD}^2_k} \coloneqq
	\frac{1}{m^2} \sum_{1\leq i, i' \leq m} k(X_i,X_{i'})
	- \frac{2}{mn} \sum_{i=1}^m \sum_{j=1}^n k(X_i,Y_j)
	+ \frac{1}{n^2}  \sum_{1\leq j, j' \leq n} k(Y_j,Y_{j'}) 
\end{equation}
which can be expressed as a two-sample (both of second order) V-statistic \citep{lee1990ustatistic}
\begin{equation}
	V_{\mathrm{MMD}^2_k}
	=\frac{1}{m^2n^2} 
	\sum_{1\leq i, i' \leq m}
	\sum_{1\leq j, j' \leq n}
	\coremmd(X_i, X_{i'}; Y_j, Y_{j'})
\end{equation}
with core function\footnote{This is more commonly referred to as a `kernel' in the litterature, we use the term `core' to avoid confusion with the positive-definite kernel $k$.} 
\begin{equation}
	\label{h}
	\coremmd(x, x'; y, y') \coloneqq k(x,x') - k(x',y) - k(x,y') + k(y,y') 
\end{equation}
for $x,x',y,y'\in\R^d$.
Writing the estimator $V_{\mathrm{MMD}^2_k}$ as a two-sample V-statistic can be theoretically appealing but we stress that it can in fact be computed in quadratic time using \Cref{mmdagg_mmdmnV}.
The V-statistic incorporates the terms $\{k(X_i,Y_i):i=1,\dots,n\}$ and, hence, is biased.
We also point out that the V-statistic is always non-negative, and taking its square root yields an estimator of the (non-squared) MMD.
The global sensitivity of this MMD statistic is studied in \citet[Lemma 5]{kim2023differentially}, which provides robustness guarantees \citep{schrab2024robust}.

Writing $Z_i = X_i$ for $i=1,\dots,m$, $Z_{m+j} = Y_j$ for $j=1,\dots,n$, and considering the $(m+n)\times (m+n)$ kernel matrix $K^{ZZ} = \big(k(Z_i,Z_j)\big)_{1\leq i,j\leq m+n}$, the MMD V-statistic can be computed as
\begin{equation}
\label{eq:mmdU_comp}
V_{\mathrm{MMD}^2_k}
= w^\top K^{ZZ} w
\end{equation}
where $w$ is a vector of size $m+n$ with $w_i = \nicefrac{1}{m}$ for $i=1,\dots,m$ and $w_{m+j} = -\nicefrac{1}{n}$ for $j=1,\dots,n$.

When the sample sizes are equal $m=n$, the estimator reduces to a one-sample second-order V-statistic
\begin{equation}
\label{eq:mmdsecond}
V_{\mathrm{MMD}^2_k} = 
\frac{1}{n^2} 
\sum_{1\leq i, i' \leq n}
\coremmd(X_i, X_{i'}; Y_i, Y_{i'}).
\end{equation}

\paragraph{MMD U-statistic.}
An unbiased estimator of the squared MMD \citep[Lemma~6]{gretton2012kernel} naturally arises from \Cref{mmdagg_eq:mmd2} as
\begin{equation}
	\label{mmdagg_mmdmnUn}
	U_{\mathrm{MMD}^2_k} \coloneqq
	\frac{1}{m(m-1)} \sum_{1\leq i\neq i' \leq m} k(X_i,X_{i'})
	- \frac{2}{mn} \sum_{i=1}^m \sum_{j=1}^n k(X_i,Y_j)
	+ \frac{1}{n(n-1)}  \sum_{1\leq j\neq j' \leq n} k(Y_j,Y_{j'}),
\end{equation}
this is actually the minimum variance unbiased MMD estimator \citep[Section 5.1.4]{serfling1980approximation}.
It can be expressed as a two-sample (both of second order) U-statistic \citep{hoeffding1992class}
\begin{equation}
	\label{mmdagg_mmdmnUnn}
	U_{\mathrm{MMD}^2_k} =
	\frac{1}{m(m-1)n(n-1)} 
	\sum_{1\leq i\neq i' \leq m}
	\sum_{1\leq j\neq j' \leq n}
	\coremmd(X_i, X_{i'}; Y_j, Y_{j'}).
\end{equation}
The MMD U-statistic expression of \Cref{mmdagg_mmdmnUn} cannot be expressed as a single vector-matrix-vector product, instead it needs to be computed as
\begin{equation}
U_{\mathrm{MMD}^2_k}
= \frac{1}{m(m-1)}\one^\top \bar{K}^{XX} \one
- \frac{2}{mn}\one^\top K^{XY} \one
+ \frac{1}{n(n-1)}\one^\top \bar{K}^{YY} \one
\end{equation}
where 
$K^{XX} = \big(k(X_i,X_j)\big)_{1\leq i,j\leq m}$,
$K^{XY} = \big(k(X_i,Y_j)\big)_{1\leq i\leq m, 1\leq j\leq n}$,
$K^{YY} = \big(k(Y_i,Y_j)\big)_{1\leq i,j\leq n}$, 
where $\bar{K}^{XX}$ and $\bar{K}^{YY}$ denote the matrices $K^{XX}$ and $K^{YY}$ with diagonal entries set to zero,
and where $\one$ is a vector of ones of size either $m$ or $n$ depending on the context.
Efficient implementations of the MMD U-statistic are further discussed in \citet{schrab2022efficient}.

When $m=n$, not incorporating the terms $\{k(X_i,Y_i):i=1,\dots,n\}$ (note that the order of the samples matters) in the statistic computation results in a simpler one-sample second-order U-statistic
\citep[Equation 4]{gretton2012kernel}
\begin{equation}
\widetilde{U}_{\mathrm{MMD}^2_k} = 
\frac{1}{n(n-1)} 
\sum_{1\leq i\neq i' \leq n}
\coremmd(X_i, X_{i'}; Y_i, Y_{i'})
\end{equation}
which can be computed as 
\begin{equation}
\widetilde{U}_{\mathrm{MMD}^2_k} 
= \frac{1}{n(n-1)}v^\top \bar{K}^{ZZ} v
\end{equation}
where $\bar{K}^{ZZ}$ is the matrix $K^{ZZ}$ as defined in \Cref{eq:mmdU_comp} but with diagonal entries set to zero, and $v\coloneqq(\one_n, -\one_n)$ is a vector of signed ones.
As a result of the U-statistic being unbiased, it is not always non-negative and hence cannot be used to estimate the MMD by taking its square root (as for the V-statistic).

\paragraph{MMD kernel choice importance.}
When using the translation-invariant kernel $\kl$, \Cref{eq:kernel_limit} implies\footnote{We use the convention that $(\cdot)_\lambda$ denotes $(\cdot)_{k_\lambda}$ throughout the thesis.} 
\begin{equation}
\begin{aligned}
	&\mathrm{MMD}_\lambda^2 \to 0 \textrm{ when } \lambda\to 0 \textrm{ or } \lambda\to\infty, \\
	&V_{\mathrm{MMD}^2_\lambda} \to \frac{1}{m} + \frac{1}{n} \textrm{ when } \lambda\to 0,
\ \textrm{ and } \
V_{\mathrm{MMD}^2_\lambda} \to 0 \textrm{ when } \lambda\to\infty, \\
	&U_{\mathrm{MMD}^2_\lambda} \to 0 \textrm{ and }\widetilde{U}_{\mathrm{MMD}^2_\lambda}\to0\ \textrm{ when }\ \lambda\to 0 \textrm{ or } \lambda\to\infty.
\end{aligned}
\end{equation}
We emphasize that this holds regardless of the relation between the distributions $P$ and $Q$, so even when $P\neq Q$ if the bandwidth $\lambda$ is not well-calibrated (in the sense that it is either too small or too large) then the estimated MMD can be very close to zero which would fail to capture the difference in distributions (even for characteristic kernel $\kl$).
This observation really highlights the importance of the choice of kernel bandwidth (and more generally of kernel) in the MMD computation.

\mysubsection{HSIC: Hilbert--Schmidt Independence Criterion}
\label{subsec:hsic}

\paragraph{HSIC measure.}
For a joint probability density $P_{XY}$ on $\mathcal{X}\times\mathcal{Y}$ with marginals $P_X$ on $\Xcal$ and $P_Y$ on $\Ycal$,
we quantify the dependence with the {\em Hilbert--Schmidt Independence Criterion} (HSIC) introduced by \citet{gretton2005kernel}, which is defined as the Hilbert--Schmidt norm of the cross-covariance operator, that is\footnote{The HSIC is most commonly defined without the square, however, we choose to define it as such for consistence with the MMD and KSD discrepancies.}
\begin{align}
    &\mathrm{HSIC}_{\hsickx,\hsicky}^2(P_{XY}) \nonumber \\
	\coloneqq\ &\|\Ccal_{P_{XY}}\|^2_{\mathrm{HS}} \nonumber\\
	=\ &\big\langle\Ccal_{P_{XY}},\Ccal_{P_{XY}}\big\rangle_{\mathrm{HS}} \nonumber\\
	=\ &\big\langle\E_{P_{XY}}\!\left[\big(\hsickx(X,\cdot) - \mu_{P_X}\big)\otimes \big(\hsicky(Y,\cdot) - \mu_{P_Y}\big)\right],\E_{P_{XY}}\!\left[\big(\hsickx(X',\cdot) - \mu_{P_X}\big)\otimes \big(\hsicky(Y',\cdot) - \mu_{P_Y}\big)\right]\big\rangle_{\mathrm{HS}} \nonumber\\
	=\ &\E_{P_{XY},P_{XY}}\!\left[\big\langle\big(\hsickx(X,\cdot) - \mu_{P_X}\big)\otimes \big(\hsicky(Y,\cdot) - \mu_{P_Y}\big),\big(\hsickx(X',\cdot) - \mu_{P_X}\big)\otimes \big(\hsicky(Y',\cdot) - \mu_{P_Y}\big)\big\rangle_{\mathrm{HS}}\right] \nonumber\\
	=\ &\E_{P_{XY},P_{XY}}\!\left[
	\big\langle
	\big(\hsickx(X,\cdot) - \mu_{P_X}\big),
	\big(\hsickx(X',\cdot) - \mu_{P_X}\big)
	\big\rangle_{\Hcal_\hsickx}
	\big\langle
	\big(\hsicky(Y,\cdot) - \mu_{P_Y}\big),
	\big(\hsicky(Y',\cdot) - \mu_{P_Y}\big)
	\big\rangle_{\Hcal_\hsicky}
	\right] \nonumber\\
	=\ &\E_{P_{XY},P_{XY}}\Big[
	\Big(
	\hsickx(X,X') 
	- \E_{X}\!\left[\hsickx(X,X')\right]
	- \E_{X'}\!\left[\hsickx(X,X')\right]
	+ \E_{X,X'}\!\left[\hsickx(X,X')\right]
	\Big) \nonumber\\
	&\phantom{\E_{P_{XY},P_{XY}}\Big[}
	\Big(
	\hsicky(Y,Y') 
	- \E_{Y}\!\left[\hsicky(Y,Y')\right]
	- \E_{Y'}\!\left[\hsicky(Y,Y')\right]
	+ \E_{Y,Y'}\!\left[\hsicky(Y,Y')\right]
	\Big)
	\Big] \nonumber\\
	=\ &\E_{P_{XY},P_{XY}}\Big[
	\hsickx(X,X') 
	\Big(
	\hsicky(Y,Y') 
	- \E_{Y}\!\left[\hsicky(Y,Y')\right]
	- \E_{Y'}\!\left[\hsicky(Y,Y')\right]
	+ \E_{Y,Y'}\!\left[\hsicky(Y,Y')\right]
	\Big)
	\Big] \nonumber\\
	=\ &\E_{P_{XY},P_{XY}}\Big[\hsickx(X,X')\hsicky(Y,Y')\Big] - 2\, \E_{P_{XY}}\Big[\E_{P_{X}}[\hsickx(X,X')]\E_{P_{Y}}[\hsicky(Y,Y')]\Big] \\
    &\hspace{5.32cm}+ \E_{P_X,P_X}\Big[\hsickx(X,X')\Big] \E_{P_Y,P_Y}\Big[\hsicky(Y,Y')\Big]\nonumber
\end{align}
for kernels $\hsickx$ and $\hsicky$ on $\Xcal\times\Xcal$ and $\Ycal\times\Ycal$, respectively, where we have used the property of the cross-covariance operator shown in \Cref{eq:Cxy_property}.
We also mention the related conditional HSIC quantities of \citet{zhang2011kernel} and \citet{pogodin2022efficient,pogodin2024practical}.

\paragraph{HSIC V-statistic.}
We now present some HSIC estimators given i.i.d.\ paired samples $\big((X_i,Y_i)\big)_{i=1}^N$ drawn from $P_{XY}$.
For convenience, we use the notation $Z_i = (X_i,Y_i)$ for $i=1,\dots,N$.
We also denote by $\widehat{P}_{XY}$ the empirical distribution of the joint.
For notational purposes, we let $\hsicKx_{ij}$ and $\hsicKy_{ij}$ denote $\hsickx(X_i, X_j)$ and $\hsicky(Y_i, Y_j)$, respectively, for all $1\leq i,j\leq N$.
The plug-in estimator \citep[Equation 4]{gretton2008kernel} of $\mathrm{HSIC}^2_{\hsickx\times\hsicky}(P_{XY})$ is $\mathrm{HSIC}^2_{\hsickx\times\hsicky}(\widehat{P}_{XY})$ which is equal to
\begin{align}
V_{\mathrm{HSIC}^2_{\hsickx\!,\hsicky}}
	&\coloneqq
    \frac{1}{N^2}\!\!
    \sum_{1\leq i,j\leq N}
	\hsicKx_{ij} \hsicKy_{ij}
    - 
    \frac{2}{N}
	\sum_{i=1}^N
	\bigg(
	\frac{1}{N}\sum_{j=1}^N \hsicKx_{ij}
	\bigg)
	\bigg(
	\frac{1}{N}\sum_{r=1}^N \hsicKy_{ir}
	\bigg) 
	+
	\bigg(\frac{1}{N^2} \!\!
    \sum_{1\leq i,j\leq N}
	\hsicKx_{ij}
	\bigg)
    \bigg(
    \frac{1}{N^2} \!\!
    \sum_{1\leq r,s\leq N}
	\hsicKy_{rs}
	\bigg)\nonumber\\[0.5cm]
	&=
    \frac{1}{N^2}
    \sum_{1\leq i,j\leq N}
    \hsicKx_{ij} \hsicKy_{ij}
    - 
    \frac{2}{N^3}
    \sum_{1\leq i,j,r\leq N}
    \hsicKx_{ij} \hsicKy_{ir} 
	+
    \frac{1}{N^4} 
    \sum_{1\leq i,j,r,s\leq N}
    \hsicKx_{ij} \hsicKy_{rs}\nonumber\\
	&=
	\label{eq:hsicVV}
    \frac{1}{N^4} 
    \sum_{1\leq i,j,r,s\leq N}
	\hsicKx_{ij}
	\Big( 
	\hsicKy_{ij} - \hsicKy_{is}- \hsicKy_{rj}+ \hsicKy_{rs}
	\Big)\\
	&=
    \frac{1}{N^4} 
    \sum_{1\leq i,j,r,s\leq N}
	\frac{1}{4}
	\Big( 
	\hsicKx_{ij} - \hsicKx_{is}- \hsicKx_{rj}+ \hsicKx_{rs}
	\Big)
	\Big( 
	\hsicKy_{ij} - \hsicKy_{is}- \hsicKy_{rj}+ \hsicKy_{rs}
	\Big).
\end{align}
So, this HSIC estimator can be expressed as a one-sample fourth-order V-statistic
\begin{equation}
    \label{eq:hsicsecond}
    V_{\mathrm{HSIC}^2_{\hsickx\!,\hsicky}}
    =
    \frac{1}{N^4} 
    \sum_{1\leq i,j,r,s\leq N}
    \corehsick(Z_i, Z_j, Z_r, Z_s)	
\end{equation}
where the core HSIC function can either be defined as
\begin{equation}
    \corehsick(Z_i, Z_j, Z_r, Z_s)	
    = \hsicKx_{ij}
	\Big( 
	\hsicKy_{ij} - \hsicKy_{is}- \hsicKy_{rj}+ \hsicKy_{rs}
	\Big)
	= \hsickx(X_i, X_j) h^{\operatorname{MMD}}_\hsicky(Y_i, Y_j; Y_r, Y_s)
\end{equation}
or
\begin{equation}
\begin{aligned}
\label{eq:hsiccore}
	\corehsick(Z_i, Z_j, Z_r, Z_s)	
	&= \frac{1}{4}
	\Big( 
	\hsicKx_{ij} - \hsicKx_{is}- \hsicKx_{rj}+ \hsicKx_{rs}
	\Big)
	\Big( 
	\hsicKy_{ij} - \hsicKy_{is}- \hsicKy_{rj}+ \hsicKy_{rs}
	\Big)\\
	&= \frac{1}{4}h^{\operatorname{MMD}}_\hsickx(X_i, X_j; X_r, X_s)h^{\operatorname{MMD}}_\hsicky(Y_i, Y_j; Y_r, Y_s),
\end{aligned}
\end{equation}
where the second expression has the benefit of being symmetric in the samples but this comes at the expense of computing four times the same quantities.
In this thesis, we will use the second expression for the core HSIC function.
Again, this estimator is non-negative and biased (as it includes the terms with same indices), its square root can be taken to estimate the (non-squared) HSIC directly, and its global sensitivity is studied in \citet[Lemma 6]{kim2023differentially}.
We stress that the HISC estimator $V_{\mathrm{HSIC}^2_{\hsickx\!,\hsicky}}$ can be computed in quadratic time and admits the following closed-form expression \citep[Equation 4]{gretton2008kernel}
\begin{equation}
    V_{\mathrm{HSIC}^2_{\hsickx\!,\hsicky}}
    =
	\frac{1}{N^2} \tr\!\left(\hsicKx H\hsicKy H\right)
\end{equation}
where $\hsicKx = \big(\hsicKx_{ij}\big)_{1\leq i,j\leq N}$, $\hsicKy = \big(\hsicKy_{ij}\big)_{1\leq i,j\leq N}$ and $H=I - \frac{1}{N}\one\one^\top$ is the centering matrix with $I$ the identity matrix and $\one$ a vector of ones, and where $\tr$ denotes the trace of a matrix.
Finally, we present a one-sample second-order V-statistic for the HSIC, which does not consider all possible terms (unlike the one presented above) but which is useful to construct efficient estimators with faster computation time, as discussed in \Cref{sec:fixed_estimators},
\begin{equation}
\label{eq:hsic_second_order}
    {\widetilde V}_{\mathrm{HSIC}^2_{\hsickx\!,\hsicky}}
    =
    \frac{1}{N^2}
    \sum_{1\leq i,j\leq N}
    \corehsick(Z_i, Z_j, Z_{i+N/2}, Z_{j+N/2})
\end{equation}
where the indices are taken modulo $N$ which is here assumed to be even. 
While shifting the indices by other quantities than $N/2$ is possible, this choice turns out to be particularly useful due to the property that adding this shift twice to an index simply leaves the index unchanged (useful in the setting of \citealp{schrab2022efficient}).

\paragraph{HSIC U-statistic.}
A natural unbiased HSIC estimator \citep{gretton2008kernel,song2012feature} is the minimum variance one-sample fourth-order U-statistic
\begin{equation}
\label{eq:Uhsic}
\begin{aligned}
U_{\mathrm{HSIC}^2_{\hsickx\!,\hsicky}}
	&\coloneqq
    \frac{1}{\abss{\mathbf{i}_2^N}}
	\sum_{(i,j)\in \mathbf{i}_2^N}
    \hsicKx_{ij} \hsicKy_{ij}
    - 
    \frac{2}{\abss{\textbf{i}_3^N}}
	\sum_{(i,j,r)\in \mathbf{i}_3^N}
    \hsicKx_{ij} \hsicKy_{ir} 
	+
    \frac{1}{\abss{\mathbf{i}_4^N}} 
	\sum_{(i,j,r,s)\in \mathbf{i}_4^N}
    \hsicKx_{ij} \hsicKy_{rs}\\
	&=
    \frac{1}{\abss{\mathbf{i}_4^N}} 
	\sum_{(i,j,r,s)\in \mathbf{i}_4^N}
    \corehsick(Z_i, Z_j, Z_r, Z_s).
\end{aligned}
\end{equation}
Here, $\mathbf{i}_r^N$ denotes the set of all $r$-tuples drawn without replacement from $\{1,\dots,N\}$ so that $\abss{\textbf{i}_r^N} = N \cdots (N-r+1)$, for example $\textbf{i}_2^N=\{(i,j):1\leq i\neq j \leq N\}$ and $\abss{\textbf{i}_2^N} = N(N-1)$.

We stress the fact that this HSIC U-statistic can actually be computed in quadratic time as shown by \citet[Equation 5]{song2012feature} who provide the following closed-form expression
\begin{equation}
    U_{\mathrm{HSIC}^2_{\hsickx\!,\hsicky}}
    = \frac{1}{N(N-3)}
    \left(
	\tr\pp{\bhsicKx\bhsicKy}
    +
    \frac
    {\one^\top\bhsicKx\one\one^\top\bhsicKy\one}
    {(N-1)(N-2)}
    -
    \frac{2}{N-2}
    \one^\top\bhsicKx\bhsicKy\one
    \right)
\end{equation}
where $\bhsicKx$ and $\bhsicKy$ are the kernel matrices $\hsicKx$ and $\hsicKy$ with diagonal entries set to 0.

\paragraph{HSIC kernel choice importance.}
For translation-invariant kernel $\km$ and $\kn$ with bandwidths $\lambda$ and $\mu$, \Cref{eq:kernel_limit} implies\footnote{We use the convention that $(\cdot)_{\lambda,\mu}$ denotes $(\cdot)_{\km\times\kn}$.} 
\begin{equation}
\begin{aligned}
	&\mathrm{HSIC}^2_{\lambda,\mu} \to 0 \textrm{ when } \lambda\to 0  \textrm{ or } \mu\to 0 \textrm{ or } \lambda\to\infty\textrm{ or } \mu\to\infty, \\
	&V_{\mathrm{HSIC}^2_{\lambda,\mu}} \to \frac{1}{N} - \frac{1}{N^2} \textrm{ when } \lambda\to 0  \textrm{ and } \mu\to 0,
\ \ \textrm{ and } \ \
V_{\mathrm{HSIC}^2_{\lambda,\mu}} \to 0 \textrm{ when } \lambda\to\infty\textrm{ or } \mu\to\infty, \\
	&U_{\mathrm{HSIC}^2_{\lambda,\mu}} \to 0 \textrm{ when } \lambda\to 0  \textrm{ or } \mu\to 0 \textrm{ or } \lambda\to\infty\textrm{ or } \mu\to\infty.
\end{aligned}
\end{equation}
Again, we stress that this holds regardless of the potential dependence in the joint distribution.
This means that even if strong dependence exists, it will be failed to be captured by the (estimated) HSIC if either of the kernel bandwidths are not well-calibrated (in the sense that they are either too small or too large).
These remarks emphasize the crucial role of kernel selection when using HSIC in practical applications.

\paragraph{HSIC as an MMD.}
First, we note that the HSIC is an MMD between the joint and the product of the marginals using a product kernel defined as $(\hsickx\times\hsicky)\big((x,y), (x',y')\big) \coloneqq \hsickx(x,x')\hsicky(y,y')$ for any $(x,y),(x',y')\in\Xcal\times\Ycal$, that is
\begin{align}
	&\mathrm{MMD}_{\hsickx\times\hsicky}^2(P_{XY}, P_X\otimes P_Y) \nonumber\\
	=\ &\E_{P_{XY},P_{XY}}[\hsickx(X,X')\hsicky(Y,Y')] - 2 \E_{P_{XY},P_XP_Y}[\hsickx(X,X')\hsicky(Y,Y')] + \E_{P_XP_Y,P_XP_Y}[\hsickx(X,X')\hsicky(Y,Y')]\nonumber\\
	=\ &\E_{P_{XY},P_{XY}}\Big[\hsickx(X,X')\hsicky(Y,Y')\Big] - 2\, \E_{P_{XY}}\Big[\E_{P_{X}}[\hsickx(X,X')]\E_{P_{Y}}[\hsicky(Y,Y')]\Big] \nonumber\\
	 &\phantom{\E_{P_{XY},P_{XY}}\Big[\hsickx(X,X')\hsicky(Y,Y')\Big]}+ \E_{P_X,P_X}\Big[\hsickx(X,X')\Big] \E_{P_Y,P_Y}\Big[\hsicky(Y,Y')\Big] \nonumber\\
	=\ &\mathrm{HSIC}^2_{\hsickx,\hsicky}(P_{XY}).
\end{align}

\paragraph{MMD as an HSIC.}
Now, consider the two-sample problem again with distributions $P,Q$, where for clarity we use variables $A,A'\iid P$ and $B,B'\sim Q$.
Construct a joint distribution $P_{XY}$ with marginal $P_X = w_P P + w_Q Q$ a mixture of $P$ and $Q$ with positive weights $w_P+w_Q=1$, and set $Y=1$ if $X$ is drawn from $P$, or $Y=-1$ if $X$ is drawn from $Q$.
For the labels, use the indicator kernel $k^\Ycal(y,y') = \one(y=y')$.
For the data, we simply set the kernel to be the one used for two-sample testing, that is $k^\Ycal(x,x') = k(x,x')$.
Then, we observe that
\begin{align*}
    \mathrm{HSIC}^2_{\hsickx,\hsicky}(P_{XY}) = \mathrm{(I)} + \mathrm{(II)} + \mathrm{(III)}
\end{align*}
where
\begin{align*}
\mathrm{(I)} 
= \E_{P_{XY},P_{XY}}\Big[\hsickx(X,X')\hsicky(Y,Y')\Big]
= w_P^2 \,\E_{P,P}\Big[k(A,A')\Big] + w_Q^2 \,\E_{Q,Q}\Big[k(B,B')\Big],
\end{align*}
and
\begin{align*}
\mathrm{(II)} &= - 2\, \E_{P_{XY}}\Big[\E_{P_{X}}[\hsickx(X,X')]\E_{P_{Y}}[\hsicky(Y,Y')]\Big]\\
&= -2 w_P^2 \left(w_P \,\E_{P,P}\Big[k(A,A')\Big] + w_Q \,\E_{P,Q}\Big[k(A,B)\Big]\right)
-2 w_Q^2 \left(w_Q \,\E_{Q,Q}\Big[k(B,B')\Big] + w_P \,\E_{P,Q}\Big[k(A,B)\Big]\right) \\
&= -2 \left(w_P^3 \,\E_{P,P}\Big[k(A,A')\Big] + w_Q^3 \,\E_{Q,Q}\Big[k(B,B')\Big] + (w_P^2 w_Q + w_P w_Q^2) \,\E_{P,Q}\Big[k(A,B)\Big]\right),
\end{align*}
and
\begin{align*}
\mathrm{(III)} &= \E_{P_X,P_X}\Big[\hsickx(X,X')\Big] \E_{P_Y,P_Y}\Big[\hsicky(Y,Y')\Big]\\
&= 
\left(w_P^2 \,\E_{P,P}\Big[k(A,A')\Big] + w_Q^2 \,\E_{Q,Q}\Big[k(B,B')\Big] + 2w_P w_Q \,\E_{P,Q}\Big[k(A,B)\Big]\right)
\left(w_P^2+w_Q^2\right).
\end{align*}
Combining these expressions, we obtain
\begin{equation}
\begin{aligned}
    \mathrm{HSIC}^2_{k,\one}(P_{XY}) &= 2 w_P^2 w_Q^2 \left(
\E_{P,P}\Big[k(A,A')\Big] - 2\,\E_{P,Q}\Big[k(A,B)\Big] + \E_{Q,Q}\Big[k(B,B')\Big]
    \right) \\
    &= 2 w_P^2 w_Q^2 \mmd^2(P,Q).
\end{aligned}
\end{equation}
In that setting, given $m$ samples from $P$ and $n$ samples from $Q$, we have $w_P=m/(m+n), w_Q=n/(m+n)$, we similarly obtain
\begin{equation}
\label{eq:hsic_mmd}
V_{\mathrm{HSIC}^2_{k,\one}}
= 
\frac{2m^2n^2}{(m+n)^4}
V_{\mathrm{MMD}^2_k}.
\end{equation}
Noting that $\kl(\cdot,\cdot)\to\one(\cdot=\cdot)$ as the bandwidth $\lambda$ shrinks to $0$, we also have
\begin{equation}
    \mathrm{HSIC}^2_{k,\kl}(P_{XY})
    \to 2 w_P^2 w_Q^2 \mmd^2(P,Q)
\qquad \textrm{ and } \qquad
V_{\mathrm{HSIC}^2_{k,\kl}}
\to 
\frac{2m^2n^2}{(m+n)^4}
V_{\mathrm{MMD}^2_k}
\end{equation}
as $\lambda\to 0$.

\mysubsection{KSD: Kernel Stein Discrepancy}
\label{subsec:ksd}

\textbf{SD measure.} 
Stein's methods \citep{stein1972bound} have been widely used in the machine learning and statistics communities (see \citet{anastasiou2021stein} for a recent review). 
At the heart of this field lies the concept of a Stein operator 
$\Acal_P\colon\Fcal\to\Gcal$
for function classes\footnote{The set $\mathrm{Func}\p{\Xcal\to\Ycal}$ consists of all functions from $\Xcal$ to $\Ycal$.} $\Fcal\subseteq \mathrm{Func}\p{\R^d\to\R^d}$ and $\Gcal\subseteq \mathrm{Func}\p{\R^d\to\R}$,
which is a linear operator satisfying Stein's identity \citep{stein1972bound,stein2004use}
\begin{equation}
    P = Q 
    \quad \Longleftrightarrow \quad
    \mathbb{E}_Q\big[(\Acal_P \fbm)(X)\big] = 0\ \textrm{ for all }\ \fbm\in\Fcal.
\end{equation}
The Stein discrepancy \citep{gorham2015measuring} is then defined as the integral probability metric (using the range of $\Acal_P$ as the function class)
\begin{equation}
	\mathrm{SD}_{\Acal_P}(P,Q)
	= \sup_{\fbm\in\mathcal{F}}\ \mathbb{E}_Q\big[(\mathcal{A}_P \fbm)(X)\big] - \mathbb{E}_P\big[(\mathcal{A}_P \fbm)(X)\big]
	=
	\sup_{\fbm\in\mathcal{F}}\ \mathbb{E}_Q\big[(\mathcal{A}_P \fbm)(X)\big].
\end{equation}
Stein operators can be constructed from infinitesimal Markov process generators.
In particular, 
assuming that the distribution $P$ admits a density $p$ with respect to the Lebesgue measure which is accessed only through the score function $\nabla\log p(x)$,
starting from the overdamped Langevin equation leads to the (overdamped) Langevin Stein operator $\mathcal{A}^\Lcal_P$ defined as 
\citep[Equation 4]{gorham2015measuring}
\begin{equation}
    (\mathcal{A}_P^\Lcal \fbm)(x) \coloneqq \fbm(x)^\top\nabla\log p(x) + \nabla^\top \fbm(x),
\end{equation}
where $\nabla^\top \fbm(x) = \sum_{i=1}^d \frac{\partial}{\partial x_i} f_i(x)$ is the divergence of $\fbm=(f_1,\dots,f_d)$ (\ie, the trace of the Jacobian matrix of $\fbm$).
The Langevin Stein operator can be expressed as a diffusion Stein operator
\citep[Section 3.1]{gorham2017measuring}
\begin{equation}
\begin{aligned}
	(\mathcal{A}_P^\Lcal \fbm)(x) =\ &\fbm(x)^\top\nabla\log p(x) + \nabla^\top \fbm(x)\\
	=\ &\fbm(x)^\top \p{\frac{\nabla p(x)}{p(x)}}+ \nabla^\top \fbm(x)\\
	=\ &\frac{1}{p(x)}\p{\fbm(x)^\top\nabla p(x) + \p{\nabla^\top \fbm(x)}p(x)}\\
	=\ &\frac{1}{p(x)}\p{\nabla^\top \big(\fbm(x)p(x)\big)}.
\end{aligned}
\end{equation}
Using this expression, we can indeed verify that the Stein's identity holds 
\begin{equation}
	\label{eq:ksd_vanish}
	\mathbb{E}_P\big[(\Acal_P^\Lcal \fbm)(X)\big]
	= \int_{\R^d} (\Acal_P^\Lcal \fbm)(x) p(x) \,\dd x
	= \int_{\R^d}  \nabla^\top \big(\fbm(x)p(x)\big)\,\dd x
	= \sum_{i=1}^d \int_{\R^d} \frac{\partial}{\partial x_i} \big(f_i(x)p(x)\big)\,\dd x
	= 0
\end{equation}
for functions $\fbm$ such that $f_i(x)p(x)$ vanishes at the boundaries of the domain for $i=1,\dots,d$.
Note also that \citep{ley2013stein}
\begin{equation}
\label{eq:ley}
	\mathbb{E}_Q\big[(\Acal_P^\Lcal \fbm)(X)\big]
	= 
	\mathbb{E}_Q\big[(\Acal_P^\Lcal \fbm)(X) - (\Acal_Q^\Lcal \fbm)(X)\big]
	=
	\mathbb{E}_Q\big[\fbm(X)^\top\big(\nabla\log p(X) - \nabla\log q(X)\big)\big].
\end{equation}

\paragraph{KSD measure.} 
We present the KSD constructions of \citet{chwialkowski2016kernel} and \citet{liu2016kernelized}, more precisely, we follow the notation of the former.
Let $\Hcal$ be an RKHS in $\mathrm{Func}\p{\R^d\to\R}$ with reproducing kernel $k$.
Denote by $\Hcal^d$ the product RKHS consisting of elements of the form $\fbm = (f_1,\dots,f_d)$ with $f_i\in\Hcal$ for $i=1,\dots,d$, with the associated inner product $\langle \fbm, \gbm \rangle_{\Hcald} = \sum_{i=1}^d \langle f_i, g_i \rangle_{\Hcal}$ for all $\fbm, \gbm\in\Hcald$.
Note that elements of $\Hcald$ can be seen as elements of $\mathrm{Func}\p{\R^d\to\R^d}$.
The aim is to express the Stein operator with $\Fcal \subseteq \Hcald$ in terms of the kernel $k$, which will then give a closed-form expression to compute the Stein discrepancy.
First, we recall the reproducing property of the kernel $k$ which implies that
(\Cref{eq:reproducing_derivative})
$$
\textstyle
f(x) = \langle f, k(x,\cdot)\rangle_{\Hcal}
\qquad
\qquad
\textrm{ and }
\qquad
\qquad
\frac{\partial}{\partial x_i}f(x) = \langle f, \frac{\partial}{\partial x_i}k(x,\cdot)\rangle_{\Hcal}
$$ 
for all $f\in\Hcal$, $x\in\R^d$ and $i=1,\dots,d$, under appropriate regularity conditions (\citealp[Theorem 1,][]{zhou2008derivative} and \citealp[Lemma C.9,][]{barp2022targeted}).
Using these properties, we can express the Stein operator in terms of the kernel $k$.
First, note that
$$
\nabla^\top \fbm(x) 
= \sum_{i=1}^d {\textstyle\frac{\partial}{\partial x_i}} f_i(x)
= \sum_{i=1}^d \langle f_i, {\textstyle\frac{\partial}{\partial x_i}} k(x,\cdot)\rangle_{\Hcal}
= \langle \fbm, \nabla k(x,\cdot)\rangle_{\Hcald}
$$
and, with the notation $\sbmp(x) \coloneqq \nabla\log p(x)$ for the score function, we have
\begin{align*}
\fbm(x)^\top\nabla\log p(x)
&= \fbm(x)^\top \sbmp(x)
= \sum_{i=1}^d f_i(x) s_i(x)
= \sum_{i=1}^d \langle f_i, k(x,\cdot) \rangle_{\Hcal}s_i(x) \\
&= \sum_{i=1}^d \langle f_i, k(x,\cdot) s_i(x)\rangle_{\Hcal}
= \langle \fbm, k(x,\cdot) \sbmp(x)\rangle_{\Hcald}\\
&= \langle \fbm, \nabla\log p(x)k(x,\cdot) \rangle_{\Hcald}.
\end{align*}
We conclude that the Stein operator can be expressed as
\begin{align}
\label{eq:stein_operator_rkhs}
	(\mathcal{A}_P^\Lcal \fbm)(x) 
	= \fbm(x)^\top\nabla\log p(x) + \nabla^\top \fbm(x) 
	= \langle \fbm, \nabla\log p(x)k(x,\cdot)+ \nabla k(x,\cdot)\rangle_{\Hcald}.
\end{align}
Writing $\xibm_P(x) \coloneqq \nabla\log p(x)k(x,\cdot)+ \nabla k(x,\cdot)$ as in \citet[Equation 1]{chwialkowski2016kernel}, by properties of mean embeddings and linearity of expectation, we obtain that
\begin{align}
	\E_Q\!\left[(\mathcal{A}_P^\Lcal \fbm)(X)\right]
	= \E_Q\!\left[\langle \fbm, \xibm_P(X)\rangle_{\Hcald}\right]
	= \big\langle \fbm, \E_Q\!\left[\xibm_P(X)\right]\big\rangle_{\Hcald}
.
\end{align}
The last equality holds under the Bochner integrability condition $\E_Q\!\left[\|\xibm_P(X)\|_{\Hcald}\right]<\infty$ which itself holds provided by $\E_Q[h_P(X,X)]<\infty$ since $\E_Q\!\left[\|\xibm_P(X)\|_{\Hcald}\right]\leq \sqrt{\E_Q\!\left[\|\xibm_P(X)\|_{\Hcald}^2\right]} = \sqrt{\E_Q[h_P(X,X)]}$ as shown by \citet[Theorem 2.1]{chwialkowski2016kernel}.
The Stein discrepancy with $\Fcal = \{\fbm\in\Hcald:\|\fbm\|_{\Hcald}\leq 1\}$, which is referred to as KSD for Kernel Stein Discrepancy, is then equal to
\begin{equation}
	\mathrm{KSD}_P(Q)
	=
	\sup_{\fbm\in\Fcal}\ \mathbb{E}_Q\big[(\mathcal{A}_P^\Lcal \fbm)(X)\big]
	=
	\sup_{\fbm\in\Fcal}\ \big\langle \fbm, \E_Q\!\left[\xibm_P(X)\right]\big\rangle_{\Hcald}
	= \big\|\E_Q\!\left[\xibm_P(X)\right]\!\big\|_{\Hcald}.
\end{equation}
Hence, the squared KSD can be expressed as
\citep[Theorem 2.1]{chwialkowski2016kernel}
\begin{equation}
\begin{aligned}
\mathrm{KSD}^2_P(Q)
&= \big\|\E_Q\!\left[\xibm_P(X)\right]\!\big\|_{\Hcald}^2
= \big\langle \E_Q\!\left[\xibm_P(X)\right], \E_Q[\xibm_P(Y)]\big\rangle_{\Hcald}\\
&\overset{(\star)}{=}\E_{Q,Q}\!\left[\langle \xibm_P(X), \xibm_P(Y)\rangle_{\Hcald}\right]
= \E_{Q,Q}[h_P(X,Y)]
\end{aligned}
\end{equation}
where the Stein kernel $h_P$ is defined as
\begin{equation}
\label{eq:h_ksd_stein}
\begin{aligned}
h_P(x,y) = \langle \xibm_P(x), \xibm_P(y)\rangle_{\Hcald} 
=\ &\left(\nabla\log p(x)^\top\nabla\log p(y)\right) k(x,y)
+ \nabla\log p(x)^\top \nabla_y k(x,y) \\
&+ \nabla\log p(y)^\top \nabla_x k(x,y)
+ \big\langle \nabla k(x,\cdot), \nabla k(y,\cdot)\big\rangle_{\Hcald}
\end{aligned}
\end{equation}
where $
\langle \nabla k(x,\cdot), \nabla k(y,\cdot)\rangle_{\Hcald} 
= \sum_{i=1}^d \big\langle \frac{\partial}{\partial x_i} k(x,\cdot), \frac{\partial}{\partial y_i} k(y,\cdot)\big\rangle_{\Hcal}
= \sum_{i=1}^d \frac{\partial^2}{\partial x_i \partial y_i}k(x,y)
$
by the reproducing property with derivatives of \Cref{eq:reproducing_derivative} (see also \citet[Lemma 4.34]{steinwart2008support}).
Again, the equation $(\star)$ holds under Bochner integrability which is satisfied when $\E_Q[h_P(X,X)]<\infty$.
Stein's identity gives $\mathbb{E}_P\!\big[(\mathcal{A}_P^\Lcal \fbm)(x)\big] = 0$ for all $\fbm\in\Fcal$, which implies that $\E_P[\xibm_P(X)] = 0$, and hence $\E_P[h_P(X,\cdot)] = 0$.

Writing $\dbm(x)=\sbmp(x)-\sbmq(x)$ for the difference in score, we recall that \Cref{eq:ley} gives that for any $\fbm\in\Fcal$ we have
\begin{equation}
\label{eq:ley2}
\E_Q\left[\fbm(X)^\top\dbm(X)\right]
=
\E_Q\left[(\mathcal{A}_P^\Lcal \fbm)(X)\right].
\end{equation}
As shown by \citet[Theorem 3.6]{liu2016kernelized}, the KSD can also be expressed as
\begin{equation}
\label{eq:ksd_delta}
\begin{aligned}
\E_{Q,Q}\left[
k(X,Y)\dbm(Y)^\top\dbm(X)
\right]&=
\E_{Q,Q}\left[
\Big(k(X,Y) \sbmp(X) + \nabla_{\!X}k(X,Y)\Big)^\top\dbm(Y)
\right]\\&=
\E_{Q,Q}\left[
h_P(X,Y)
\right]\\&=
\mathrm{KSD}^2_P(Q)
\end{aligned}
\end{equation}
where the first and second equalities hold by \Cref{eq:ley2} with $\fbm_1(X)=k(X,Y)\dbm(Y)$ for fixed $Y$ and $\fbm_2(Y)=k(X,Y) \sbmp(X) + \nabla_{\!X}k(X,Y)$ for fixed $X$, respectively, giving
\begin{align*}
(\mathcal{A}_P^\Lcal \fbm_1)(X)
&= k(X,Y) \dbm(Y)^\top \sbmp(X) + \nabla_{\!X}^\top\Big(k(X,Y)\dbm(Y)\Big)\\
&= \Big(k(X,Y) \sbmp(X) + \nabla_{\!X}k(X,Y)\Big)^\top\dbm(Y)
\end{align*}
as $\nabla_{\!X}^\top\big(k(X,Y)\dbm(Y)\big) = \sum_{i=1}^d \frac{\partial}{\partial X_i}k(X,Y)\delta_i(Y) = \big(\nabla_{\!X}k(X,Y)\big)^\top\dbm(Y)$,
and
\begin{align*}
(\mathcal{A}_P^\Lcal \fbm_2)(Y)
=\
&\Big(
k(X,Y) \sbmp(X) + \nabla_{\!X}k(X,Y)
\Big)^\top
\sbmp(Y)
+
\nabla_{\!Y}^\top
\Big(
k(X,Y) \sbmp(X) + \nabla_{\!X}k(X,Y)
\Big)\\=\
&k(X,Y)\sbmp(X)^\top\sbmp(Y)
+ \big(\nabla_{\!X}k(X,Y)\big)^\top  \sbmp(Y)
+ \big(\nabla_{\!Y}k(X,Y)\big)^\top  \sbmp(X)
+ \nabla_{\!Y}^\top \big(\nabla_{\!X}k(X,Y)\big)\\=\
&h_P(X,Y)
\end{align*}
as $\nabla_{\!Y}^\top \big(\nabla_{\!X}k(X,Y)\big) = \sum_{i=1}^d \frac{\partial}{\partial Y_i}\frac{\partial}{\partial X_i}k(X,Y) = \langle \nabla k(X,\cdot), \nabla k(Y,\cdot)\rangle_{\Hcald}$.
Using the Cauchy--Schwarz inequality as in \citet{liu2016kernelized}, the KSD can be upper bounded by the Fisher divergence as
\begin{equation}
\begin{aligned}
\mathrm{KSD}^2_P(Q)
&= 
\E_{Q,Q}\!\left[
k(X,Y)\dbm(X)^\top\dbm(Y)
\right]\\
&\leq
\sqrt{
\E_{Q,Q}\!\left[
k(X,Y)^2
\right] 
\E_{Q,Q}\!\left[
\big(\dbm(X)^\top\dbm(Y)\big)^2
\right]
}\\
&\leq
\sqrt{
\E_{Q,Q}\!\left[
k(X,Y)^2
\right] 
\E_{Q,Q}\!\left[
\|\dbm(X)\|_2^2 \, \|\dbm(Y)\|_2^2
\right]
}\\
&=
\sqrt{
\E_{Q,Q}\!\left[
k(X,Y)^2
\right] 
}\,
\E_{Q}\!\left[
\|\dbm(X)\|_2^2 
\right]\\
&=
\sqrt{
\E_{Q,Q}\!\left[
k(X,Y)^2
\right]}\, 
\mathrm{Fisher}(P,Q)
\end{aligned}
\end{equation}
where the Fisher divergence \citep{johnson2004information} is
\begin{equation}
\mathrm{Fisher}(P,Q)
\coloneqq 
\E_{Q}\!\left[
\|\nabla \log p(X) - \nabla \log q(X)\|_2^2
\right]. 
\end{equation}

The definition of the Stein operator $\mathcal{A}_P^\Lcal$ naturally extends to matrices $\bm{F} = (\fbm^{(1)},\dots,\fbm^{(d)})$ where each $\fbm^{(i)} = (\fbm^{(i)}_1,\dots,\fbm^{(i)}_d)$ is a vector for $i=1,\dots,d$, as
\begin{equation}
\label{eq:stein_operator_matrix}
\mathbfcal{A}_P^\Lcal \bm{F}
= \left(
\mathcal{A}_P^\Lcal \fbm^{(1)}
,\dots,
\mathcal{A}_P^\Lcal \fbm^{(d)}
\right)
\end{equation}
mapping functions from $\mathrm{Func}\p{\R^{d}\to\R^{d\times d}}$ to $\mathrm{Func}\p{\R^d\to\R^d}$.
Let $\bm{K}(x,y) = k(x,y) I_{d\times d}$ which is equal to $\big(\kbm^{(1)}(x,y),\dots,\kbm^{(d)}(x,y)\big)$ where $\kbm^{(i)}(x,y)$ is $d$-dimensional vector of zeros with $i$-th entry $k(x,y)$, giving
\begin{equation}
\label{eq:stein_operator_matrix2}
(\mathbfcal{A}_{P,x}^\Lcal \bm{K})_i (x,y)
=
\kbm^{(i)}(x,y)^\top \sbm_P(x) + \nabla^\top \kbm^{(i)}(x,y)
= k(x,y) \sbm_P(x_i) + \frac{\partial}{\partial x_i} k(x,y)
\end{equation}
for $i=1,\dots,d$, that is
\begin{equation}
\label{eq:stein_operator_matrix3}
(\mathbfcal{A}_{P,x}^\Lcal \bm{K})(x,y)
= k(x,y) \sbm_P(x) + \nabla_{\!x} k(x,y)
\end{equation}
where the subscripts $x,y$ are used to specify which variable the Stein and gradient operators are operating on.
Hence, the Stein kernel can be expressed as
\begin{equation}
\label{eq:stein_kernel_operator}
h_P(x,y) =
(\mathcal{A}_{P,y}^\Lcal\,\mathbfcal{A}_{P,x}^\Lcal\, \bm{K})(x,y)
\end{equation}
as shown above with $\mathcal{A}_P^\Lcal \fbm_2$.

\paragraph{KSD V-statistic and U-statistic.} 
We now present KSD estimators given a model distribution $P$ and some samples $X_1,\dots,X_n\iid Q$, we denote the empirical distribution by $\widehat Q$.
The Kernel Stein Discrepancy $\mathrm{KSD}_P^2(Q)$ can be estimated using a one-sample second-order U- or V-statistic
\citep{chwialkowski2016kernel,liu2016kernelized}
\begin{equation}
\label{eq:ksdsecond}
	V_{\mathrm{KSD}^2_k} \coloneqq
	\frac{1}{n^2} \sum_{1\leq i, i' \leq n} h_P(X_i,X_{i'})
	\qquad\
	\textrm{ and }
	\qquad\
	U_{\mathrm{KSD}^2_k} \coloneqq
	\frac{1}{n(n-1)} \sum_{1\leq i\neq i' \leq n} h_P(X_i,X_{i'})
\end{equation}
which can be computed as
\begin{equation}
	V_{\mathrm{KSD}^2_k} \coloneqq
	\frac{1}{n^2} \one^\top \!H^{XX} \one
	\qquad\
	\textrm{ and }
	\qquad\
	U_{\mathrm{KSD}^2_k} \coloneqq
	\frac{1}{n(n-1)} \one^\top \!H^{XX} \one
\end{equation}
where 
$H^{XX} = \big(h_P(X_i,X_j)\big)_{1\leq i,j\leq n}$ with the Stein kernel $h_P$ as in \Cref{eq:h_ksd_stein},
where $\bar{H}^{XX}$ is the matrix $H^{XX}$ with diagonal entries set to zero,
and where $\one$ is a vector of ones of size $n$.
The V-statistic corresponds to the plugin estimator $\mathrm{KSD}_P^2(\widehat Q)$ which is strictly positive and can hence be used as an estimator for the (non-squared) KSD by taking its square root, unlike the U-statistic which is unbiased but can be negative.
For consistency, the core KSD function $h_k^{\operatorname{KSD}}$ can be defined as $h_P$ itself.

\paragraph{KSD kernel choice importance.}
The behaviour of the KSD when using a translation-invariant kernel $\kl$, and letting the bandwidth $\lambda$ tend to zero, is
\begin{equation}
\mathrm{KSD}_\lambda^2
=
\E_{Q,Q}\!\left[
\kl(X,Y)\dbm(X)^\top\dbm(Y)
\right]
\to
\E_{Q}\!\left[
\dbm(X)^\top\dbm(X)
\right]
=
\mathrm{Fisher}(P,Q)^2,
\end{equation}
while, when the bandwidth $\lambda$ tends to $\infty$, we have
\begin{equation}
\mathrm{KSD}_\lambda^2
=
\E_{Q,Q}\!\left[
\kl(X,Y)\dbm(X)^\top\dbm(Y)
\right]
\to
\E_{Q,Q}\!\left[
\dbm(X)^\top\dbm(Y)
\right]
= \|
\E_{Q}\!\left[
\dbm(X)
\right]
\|_2^2.
\end{equation}
So, the KSD tends to the Fisher divergence when $\lambda\to0$, and to the 2-norm of the expected difference in score when $\lambda\to\infty$, while the MMD and HSIC tend to zero in both of these regimes.
The behaviour of the KSD U-stastistic and V-statistic when varying the bandwidth is difficult to characterise due to the complexity of the Stein kernel $h_P$ involving kernel derivatives which scale with the bandwidth.
This is drastically different from the MMD and HSIC cases, and highlights even more the importance of the bandwidth choice for the KSD statistic computation in order to obtain meaningful results.

\paragraph{Fisher divergence as a KSD.}
With the kernel $k(x,y)=\one(x=y)$, the KSD is equal to the Fisher divergence
\begin{equation}
\mathrm{KSD}_\lambda^2
=
\E_{Q,Q}\!\left[
\kl(X,Y)\dbm(X)^\top\dbm(Y)
\right]
=
\E_{Q}\!\left[
\dbm(X)^\top\dbm(X)
\right]
=
\mathrm{Fisher}(P,Q)^2.
\end{equation}
From another point of view, the KSD can be seen as a kernelized Fisher divergence.

\paragraph{KSD as an MMD.}
The MMD with the Stein kernel $h_P$ is equal to the KSD since
\begin{align}
\operatorname{MMD}_{h_P}^2(P,Q)
= \E_{Q,Q}[h_P(X,X')] - 2\,\E_{Q,P}[h_P(X,Y)] + \E_{P,P'}[h_P(Y,Y')]
= \E_{Q,Q}[h_P(X,X')] 
=\mathrm{KSD}^2_P(Q)
\end{align}
using Stein's identity $\E_P[h_P(X,\cdot)] = 0$.
We stress that the Stein kernel used in this MMD depends on the model distribution $P$.

\paragraph{MMD as a KSD.}
A simple operator can be defined as 
\begin{equation}
\label{eq:simple_stein}
(\Acal_P' \fbm)(x) = f_1(x) - \E_P[f_1(X)]
\end{equation}
for all $x\in\R^d$ and for $\fbm = (f_1,\dots,f_d)$ where $f_i\colon\R^d\to\R$ for $i=1,\dots,d$.
This is a Stein operator since $\E_P[(\Acal_P' \fbm)(X)] = 0$ for all $\fbm$.
Hence, we can define a Stein Discrepancy using this Stein operator, which we kernelise using an RKHS $\Hcal$ with reproducing kernel $k$ and unit ball $\Fcal = \{\fbm\in\Hcald:\|\fbm\|_{\Hcald}\leq 1\}$,, as
\begin{equation}
\mathrm{KSD}_{\Acal_P'}(P,Q)
~=~ \sup_{\fbm\in\Fcal}\ \E_Q\!\left[(\Acal_P' \fbm)(X)\right]
~=~ \sup_{f_1\in\Hcal\,:\,\|f_1\|_{\Hcal}\leq 1}\ \E_Q\!\left[f_1\right] - \E_P\!\left[f_1\right]
~=~ \mathrm{MMD}_k(P,Q),
\end{equation}
so the MMD itself can be seen as a KSD using a specific Stein operator.

\mysection{Efficient kernel discrepancies estimators}
\label{sec:fixed_estimators}

\paragraph{Expectation.} 
As seen in \Cref{eq:mmdsecond,eq:hsicsecond,eq:ksdsecond}, MMD, HSIC and KSD can be estimated using one-sample second-order V-statistics, which are estimators of the quantity
\begin{equation}
	\Eb{h(X,X')}
\end{equation}
for some core function $h$, and where the expectation is over independent copies $X$ and $X'$.

\paragraph{Statistics.} 
Given i.i.d. variables $X_1,\dots, X_n$, a class of estimators for this expected quantity takes the form
\begin{equation}
	\frac{1}{|\Dcal|}\sum_{(i,j) \in \Dcal} h(X_i, X_j)
\end{equation}
for some subset \(\Dcal \subseteq \cb{(i,j):1\leq i,j \leq n}\), often called the design, and can be computed in time $\bigO{|\Dcal|}$, where \(|\Dcal|\) denotes the cardinality of \(\Dcal\).

\paragraph{V-statistic.} 
The V-statistic \citep{mises1947asymptotic} is defined by setting \(\Dcal = \cb{(i,j): 1\leq i,j \leq n}\) giving 
\begin{equation}
	V = \frac{1}{n^2}\sum_{1\leq i, j \leq n} h(X_i, X_j)
\end{equation}
which can be computed in quadratic time $\bigO{n^2}$. 
Since the expectation is over independent copies, and that the V-statistic includes the terms $\cb{h(X_i,X_i):i=1,\dots,n}$, the V-statistic is biased. 

\begin{figure}[h]
	\center\includegraphics[height=5cm]{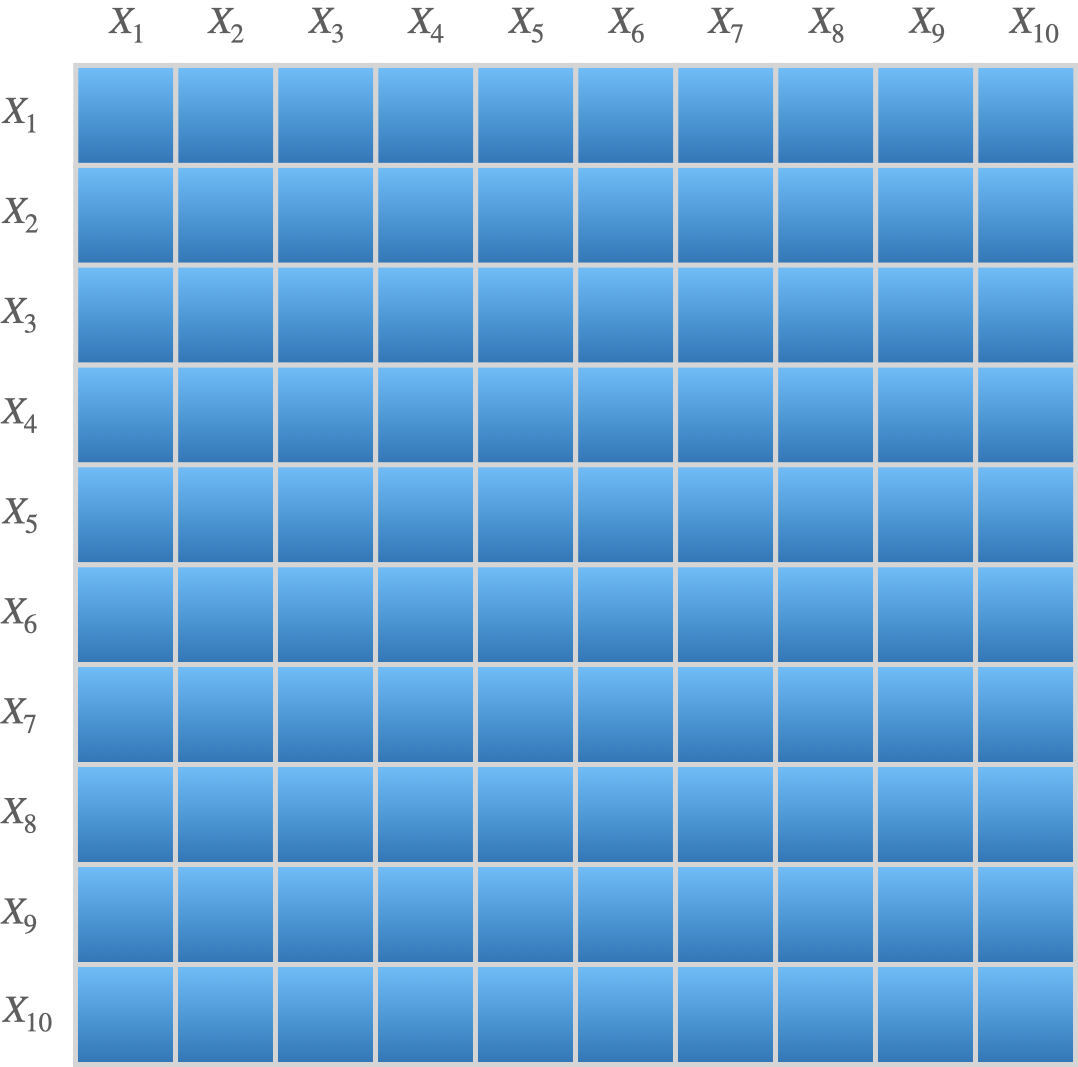}
	\caption{V-statistic. Visualisation of the core kernel matrix entries $h(X_i,X_j)$ considered (in  blue) and ignored (in white) in the sum for the V-statistic computation with $n=10$.}
\end{figure}

\paragraph{U-statistic.} 
By not including these terms, \ie by considering \(\Dcal = \cb{(i,j): 1\leq i\neq j \leq n}\), we obtain the unbiased U-statistic
\citep{hoeffding1992class,lee1990ustatistic}
\begin{equation}
	U = \frac{1}{n(n-1)}\sum_{1\leq i\neq j \leq n} h(X_i, X_j),
\end{equation}
also computable in quadratic time $\bigO{n^2}$. The U-statistic is known to be the minimum variance estimator of \(\Eb{h(X,X')}\).

\begin{figure}[h]
	\center\includegraphics[height=5cm]{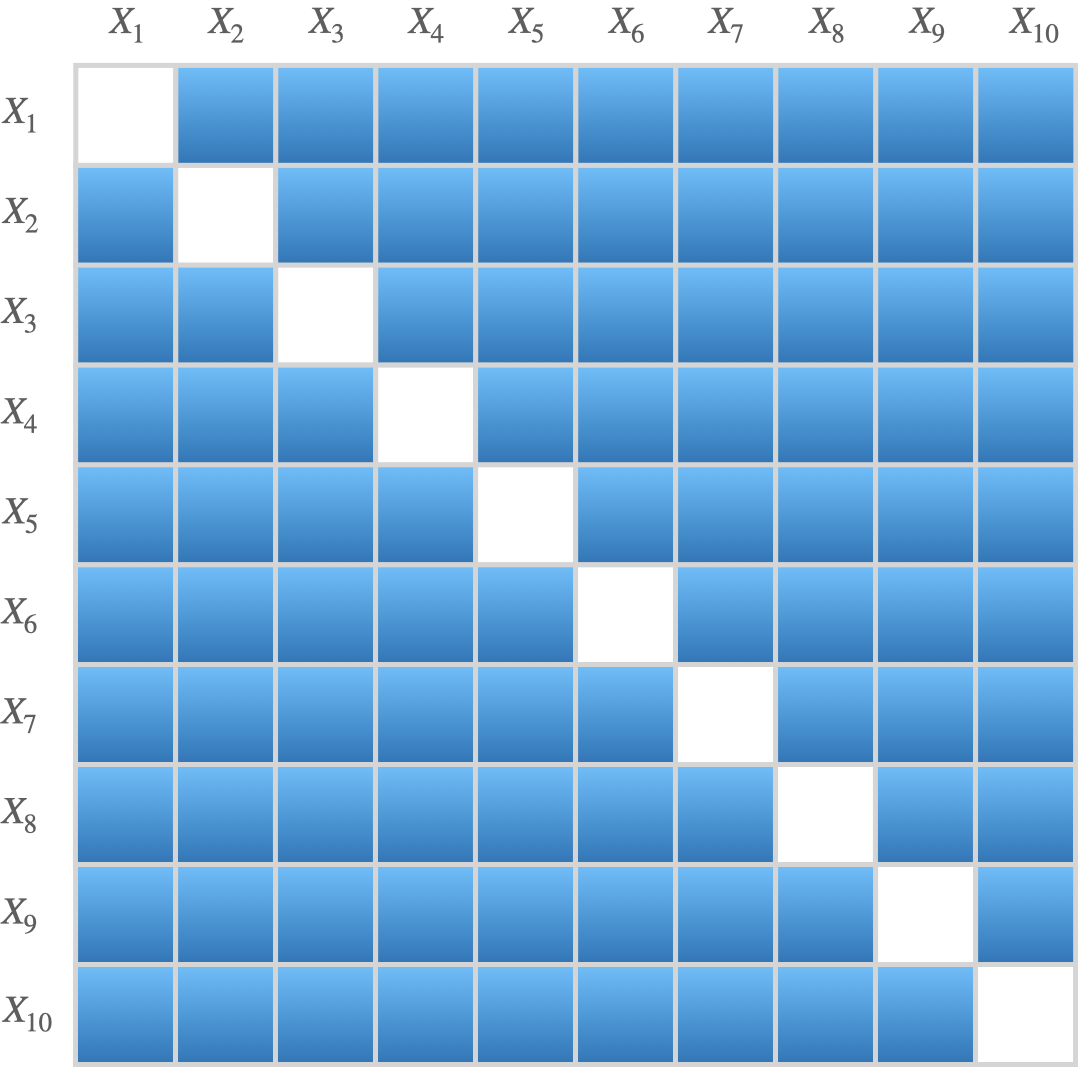}
	\caption{U-statistic. Visualisation of the core kernel matrix entries $h(X_i,X_j)$ considered (in  blue) and ignored (in white) in the sum for the U-statistic computation with $n=10$.}
\end{figure}

\paragraph{Incomplete statistic.}
The U-statistic and V-statistic are referred to as \emph{complete}, unlike their \emph{incomplete} counterparts which trade accuracy for computational efficiency \citep{blom1976some,janson1984asymptotic,lee1990ustatistic} and take the form
\begin{equation}
\frac{1}{|\Dcal|}\sum_{(i,j) \in \Dcal} h(X_i, X_j)
\end{equation}
for some strictly smaller subset \(\Dcal \subset \cb{(i,j):1\leq i\neq j \leq n}\), and can be computed in time $\bigO{|\Dcal|}$ which can be much faster than quadratic time.
Incomplete statistics are unbiased, they are particularly useful when the number of samples is large, and the kernel function is computationally expensive to evaluate.
The L-statistic, D-statistic, B-statistic, X-statistic and R-statistic, all introduced below, are examples of incomplete statistics.

Depending on the statistic, we sometimes define $\Dcal$ as a subset of the upper triangular matrix entries \(\cb{(i,j):1\leq i< j \leq n}\) and leverage the fact that the core $h$ is symmetric. 
Nonethess, in the figures, we always provide illustrations considering the full core kernel matrix.

\paragraph{L-statistic.} 
The linear L-statistic \citep[Lemma 14]{gretton2012kernel} is defined by considering the subset of the core kernel matrix entries \(\Dcal = \cb{(2i-1,2i): 1\leq i \leq \lfloor n/2 \rfloor}\), giving
\begin{equation}
	L = \frac{1}{\lfloor n/2 \rfloor}\sum_{1\leq i \leq \lfloor n/2 \rfloor} h(X_{2i}, X_{2i-1}).
\end{equation}
While this statistic can be computed in linear time $\bigO{n}$, it is rarely useful in practice as only very little information is captured when considering so few entries of the core kernel matrix.

\begin{figure}[H]
	\center\includegraphics[height=5cm]{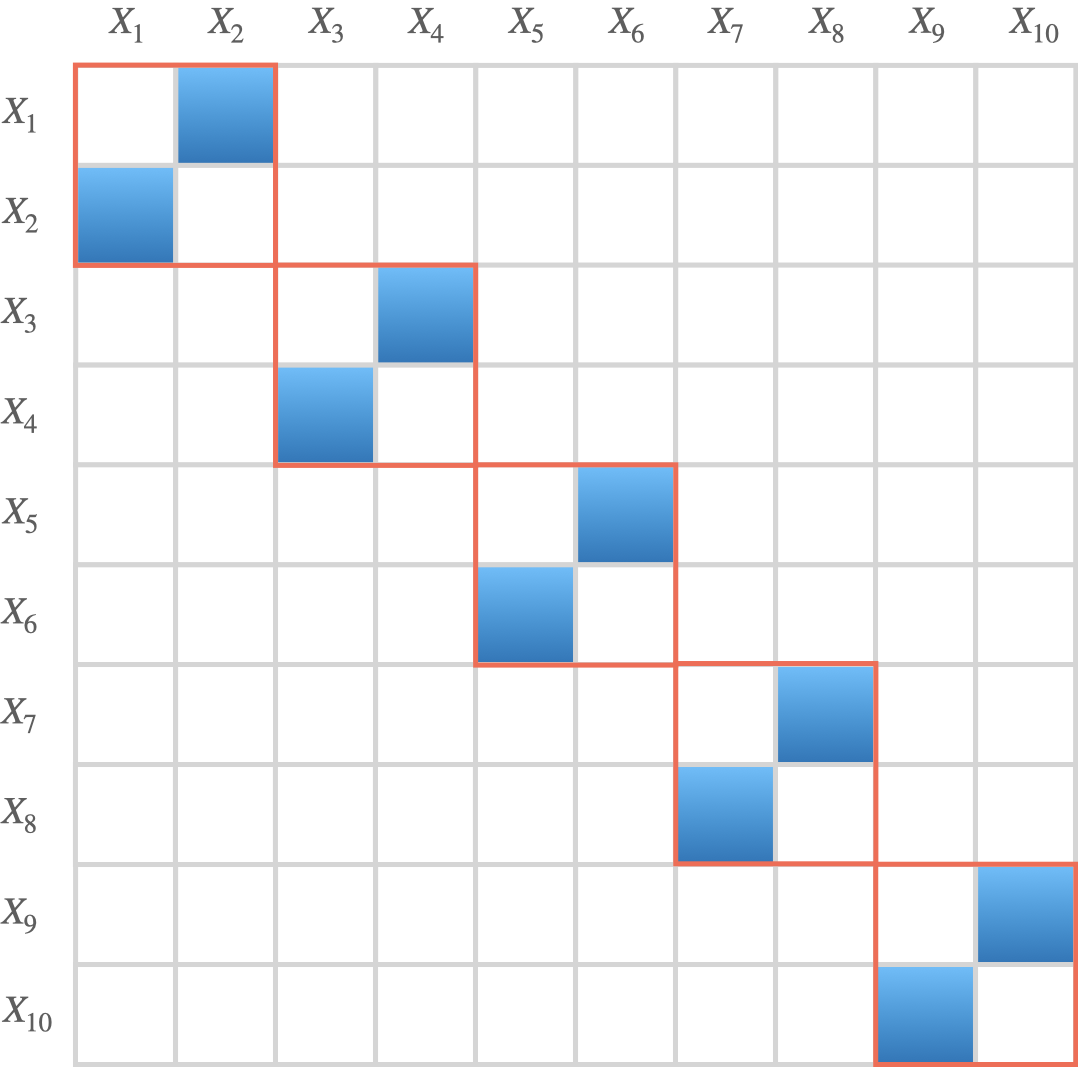}
	\caption{L-statistic. Visualisation of the core kernel matrix entries $h(X_i,X_j)$ considered (in  blue) and ignored (in white) for the L-statistic computation with $n=10$.}
\end{figure}

\paragraph{D-statistic.}
Another possibility is to include multiple subdiagonals of the core kernel matrix as done in 
\citet{schrab2022efficient}.
Considering the first $r$ subdiagonals, that is 
$\D \coloneqq \{(i,i+j): i = 1, \dots, n-j \text{ for } j = 1, \dots, r\}$ with size
$\abs{\D} = rn - r(r+1)/2$,
 gives rise to a D-statistic 
\begin{equation}
D = \frac{2}{r(2n - r-1)}\sum_{j=1}^r\sum_{i=1}^{n-j} h(X_{i}, X_{i+j}).
\end{equation}
Its time complexity is $\bigO{rn}$, if $r$ is set to a small fixed constant then this is linear, another common choice would be to set $r = \floor{\sqrt{n}}$ to obtain an estimator computable in time $\bigO{n^{1.5}}$.

\begin{figure}[H]
	\centerline{
	\includegraphics[height=5cm]{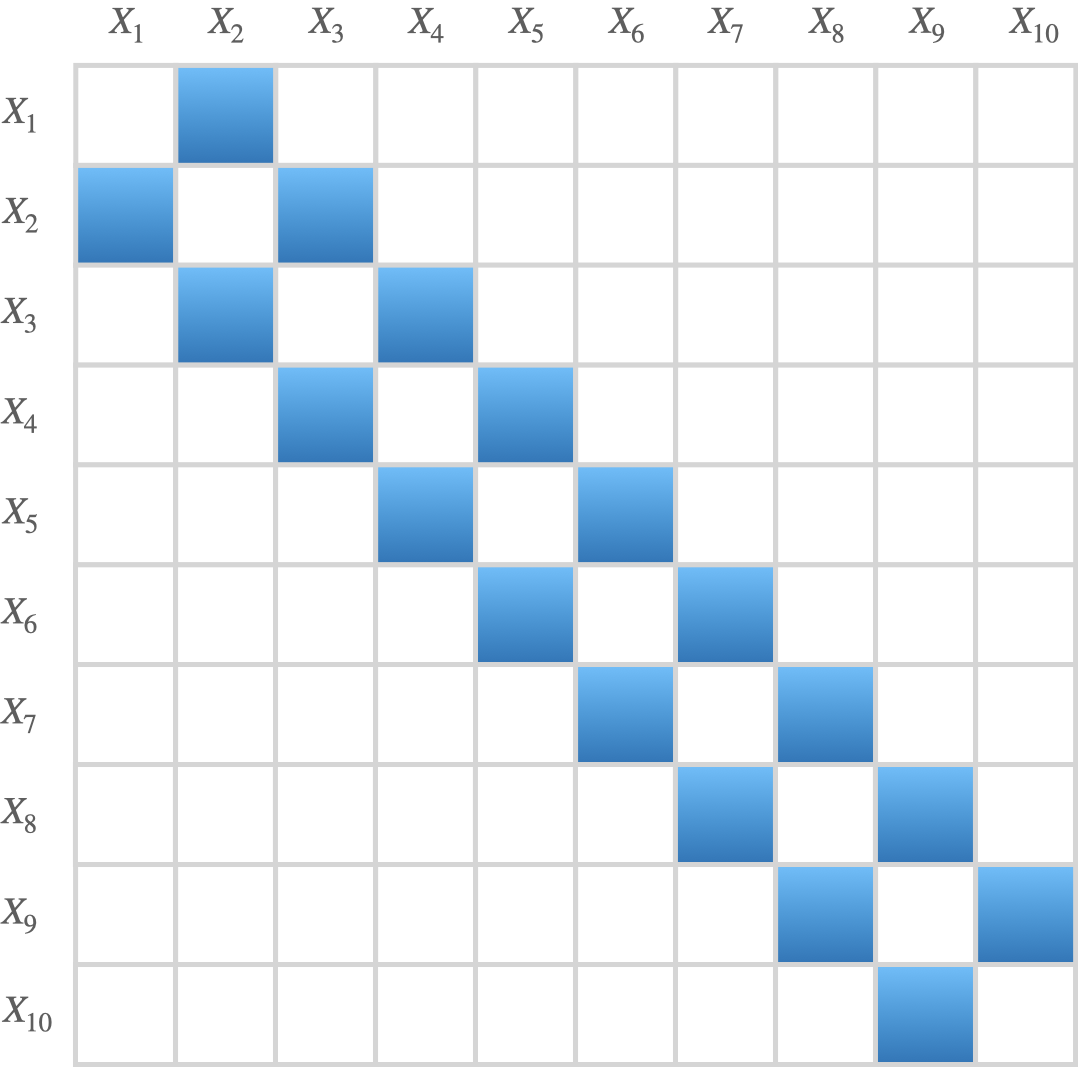}
	\hspace{0.5cm}
	\includegraphics[height=5cm]{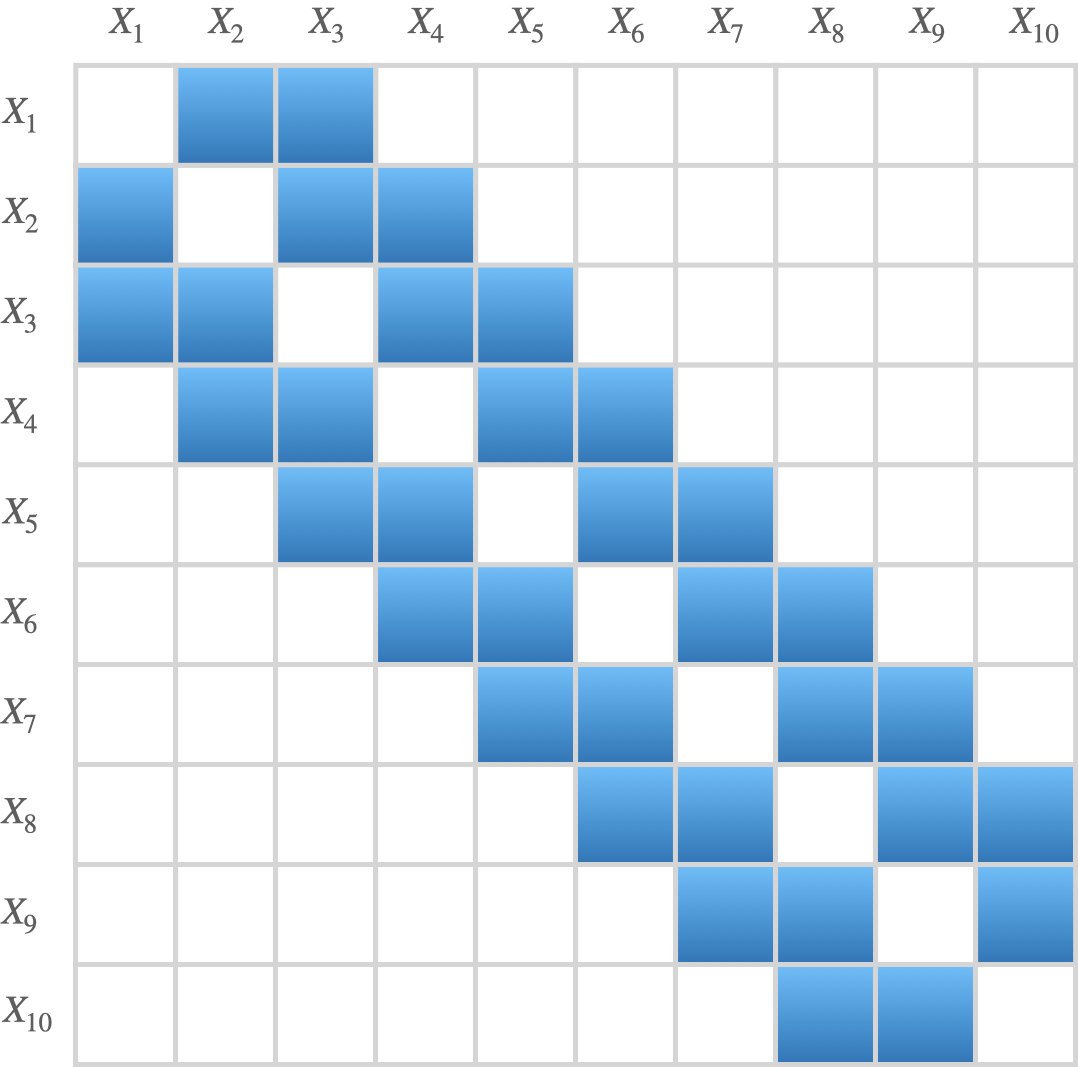}
	\hspace{0.5cm}
	\includegraphics[height=5cm]{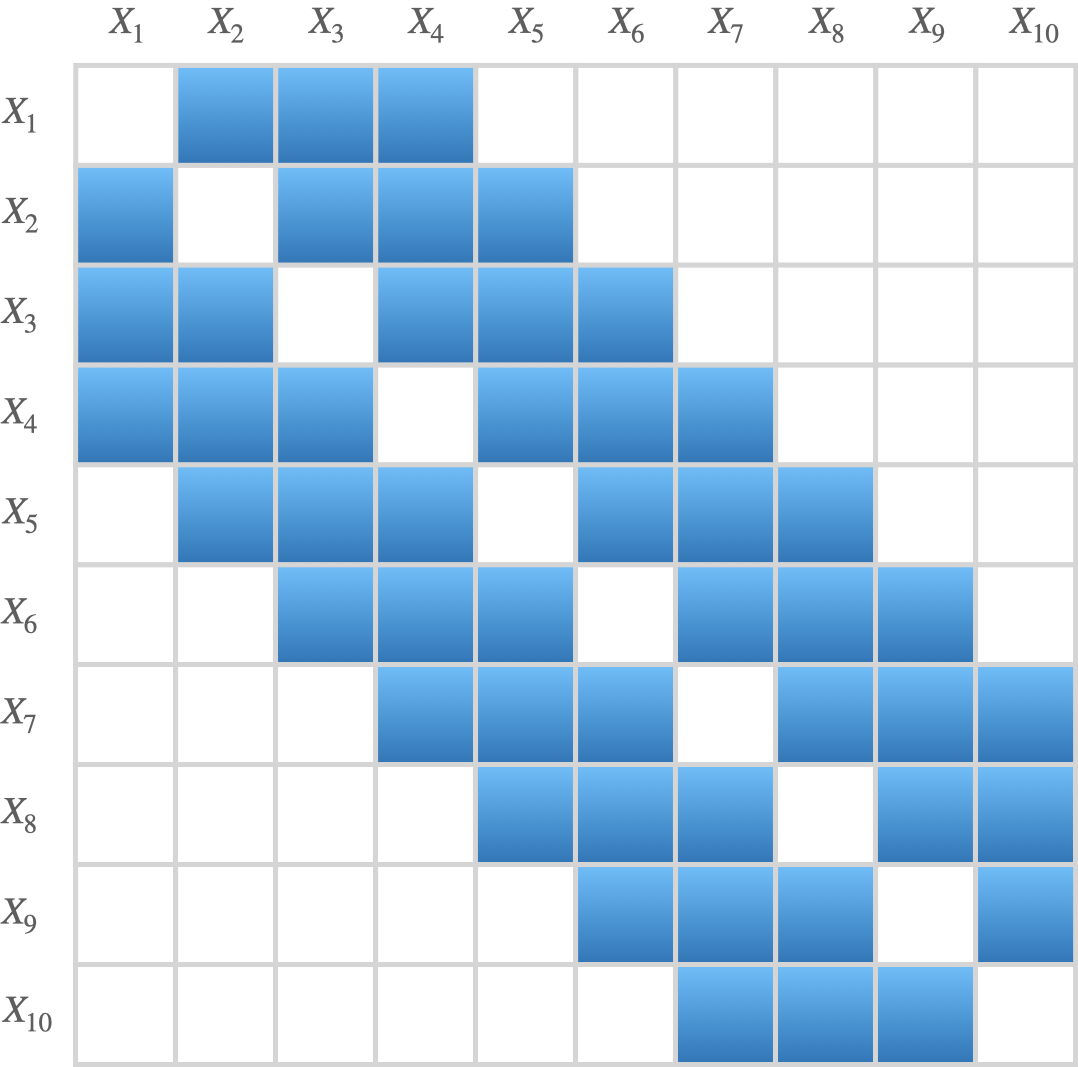}
	}
	\caption{D-statistic. Visualisation of the core kernel matrix entries $h(X_i,X_j)$ considered (in  blue) and ignored (in white) in the sum for the D-statistic computation with $n=10$.
	\emph{(Left)} $r=1$. 
	\emph{(Centre)} $r=2$. 
	\emph{(Right)} $r=3$. 
	}
\end{figure}

\paragraph{B-statistic (case $b=2$).} 
We now introduce block statistics \citep{ho2006two}, also referred to as B-statistics. 
For illustration purposes, we start with the case of $b=2$ blocks and consider some $n_1+n_2=n$.
Then, letting
\(\Dcal = \cb{(i,j): 1\leq i \neq j \leq n_1} \cup \cb{(i,j): n_1+1\leq i \neq j \leq n}\),
we obtain
\begin{equation}
\begin{aligned}
	B &= U(X_1,\dots,X_{n_1}) + U(X_{n_1+1},\dots,X_{n}) \\
	&= \frac{1}{n_1(n_1-1)}\sum_{1\leq i\neq j \leq n_1} h(X_i, X_j)
	+ \frac{1}{n_2(n_2-1)}\sum_{1\leq i\neq j \leq n_2} h(X_{n_1+i}, X_{n_1+j})
\end{aligned}
\end{equation}
with time complexity \(\bigO{n_1^2+n_2^2}\). 
It is common to consider blocks of the same size, assuming $n$ is even let $n_1 = n_2 = n/2$, then the time complexity becomes \(2 (n/2)^2\).
In this example, the block statistic is composed of U-statistics, it can also be defined similarly using V-statistics instead (in which case it would lead to a biased statistic).

\paragraph{B-statistic (general case).} 
We now consider the general case of $b$ blocks of sizes $n_1,\dots,n_b$ where \(\sum_{t=1}^b n_t = n\), and we let \(n_0=0\).
Then, considering
\begin{equation}
	\Dcal = \bigcup_{s=1}^b \cb{(i,j): 1+\sum_{t=0}^{s-1}n_t\leq i\neq j \leq \sum_{t=0}^{s}n_t} 
\end{equation}
gives the B-statistic \citep{ho2006two}
\begin{equation}
\label{eq:bstatold}
B= \frac{1}{|\Dcal|}\sum_{s=1}^b n_t(n_t-1)U\pp{X_{1+\sum_{t=0}^{s-1}n_t},\dots, X_{\sum_{t=0}^{s}n_t}}
\end{equation}
where $n_t(n_t-1)U\pp{X_{1+\sum_{t=0}^{s-1}n_t},\dots, X_{\sum_{t=0}^{s}n_t}}$ is an unscaled U-statistic, and where $|\Dcal|=\sum_{s=1}^b n_t(n_t-1)$.
This B-statistic has time complexity \(\bigO{n_1^2+\dots+n_b^2}\).
Assuming $n$ is divisible by $b$ and considering blocks of equal size $n_t=n/b$ for $t=1,\dots,b$, we obtain 
\begin{equation}
\label{eq:bstat}
B= \frac{1}{b}\sum_{s=1}^b U\pp{X_{1+(s-1)n/b},\dots, X_{sn/b}}
\end{equation}
with time complexity \(\bigO{b (n/b)^2} = \bigO{n^2/b}\). 
In practice, it is common to set $b=\floor{\sqrt{n}}$ \citep{zaremba2013b,zhang2018large} and get an estimator with time complexity of the order $\bigO{n^{1.5}}$.
When the sample size $n$ is not divisible by the number of blocks $b$, we either have one block of size strictly less than $n$ or even ignore that smaller block for simplicity.

\begin{figure}[H]
	\centerline{
	\includegraphics[height=5cm]{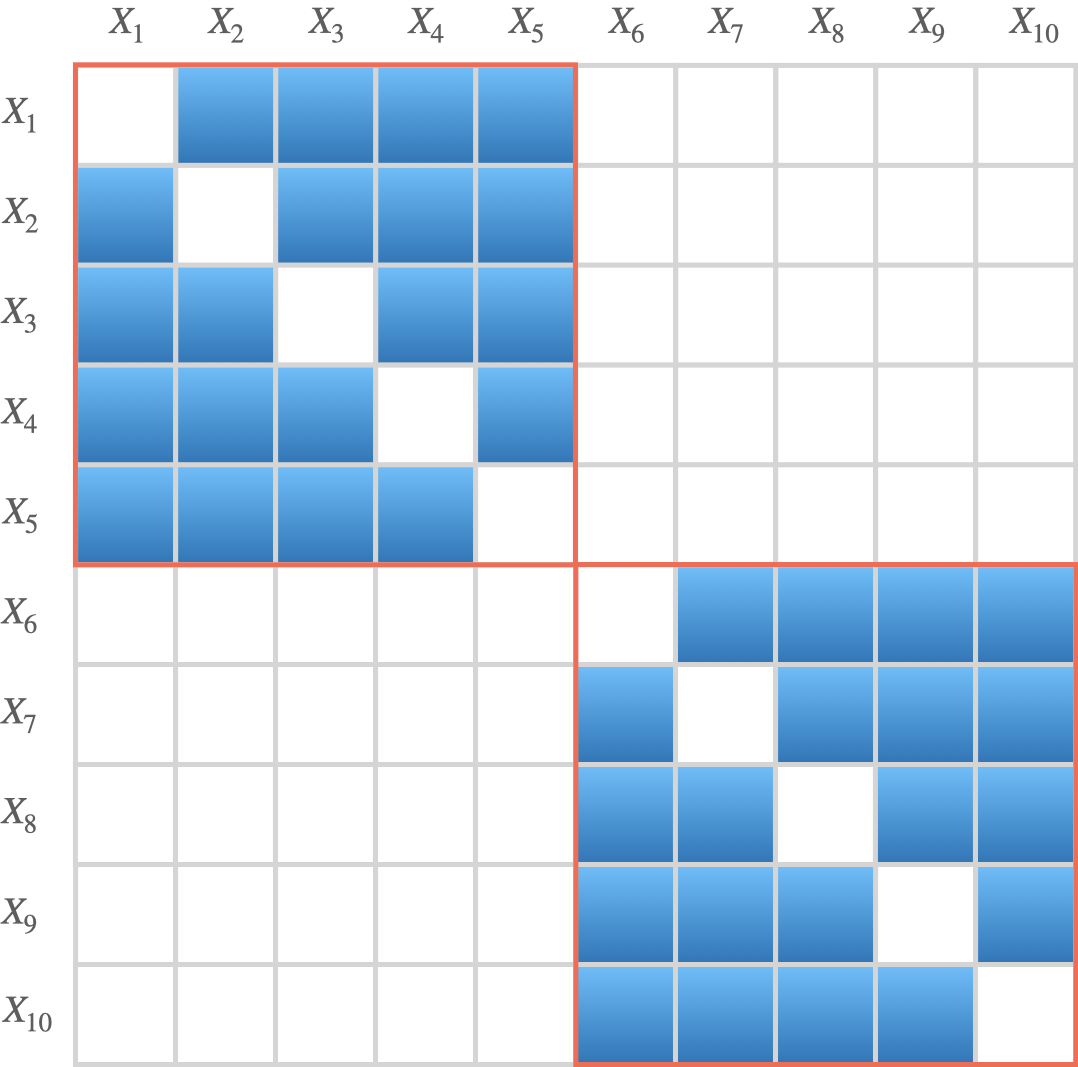}
	\hspace{0.5cm}
	\includegraphics[height=5cm]{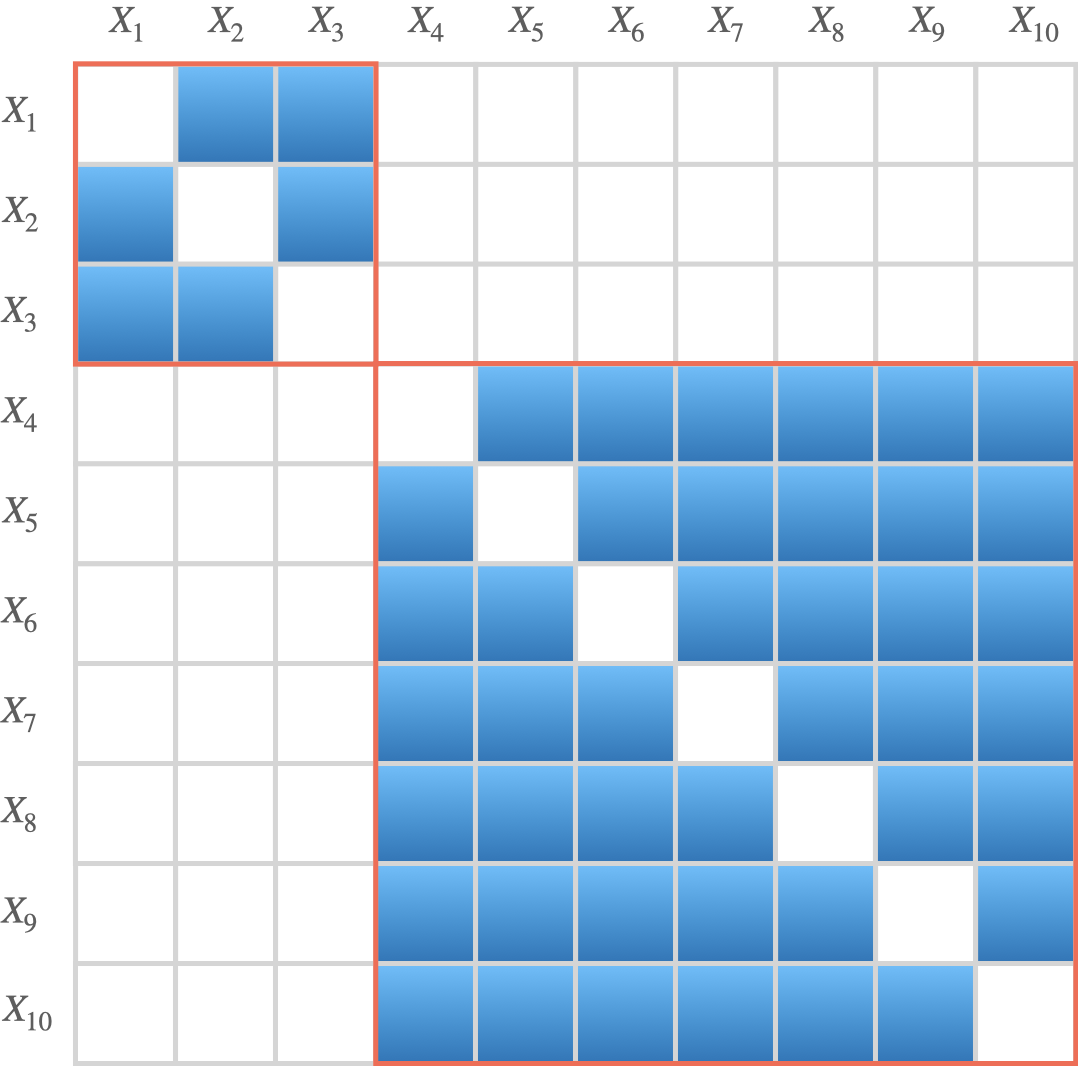}
	\hspace{0.5cm}
	\includegraphics[height=5cm]{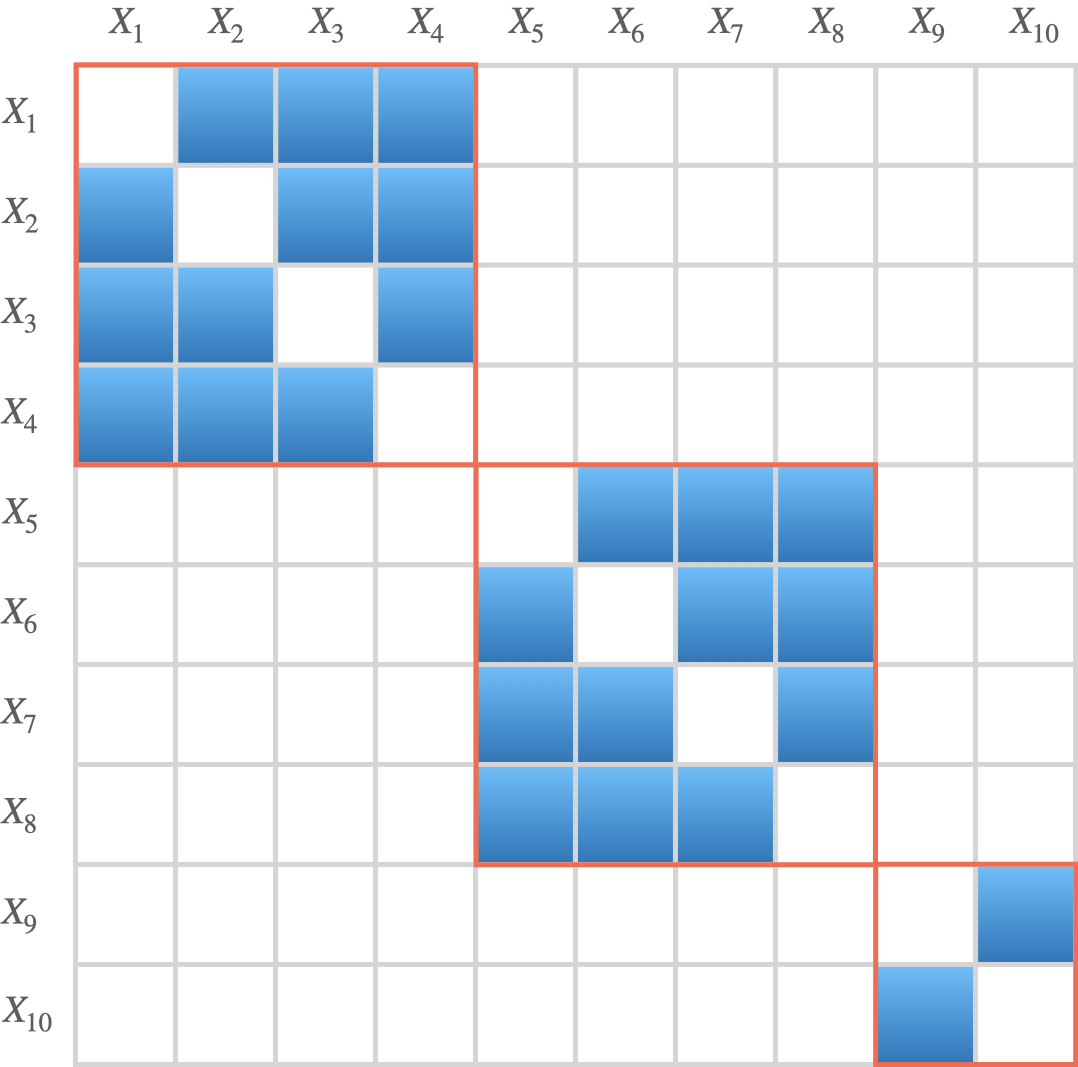}
	}
	\caption{B-statistic. Visualisation of the core kernel matrix entries $h(X_i,X_j)$ considered (in  blue) and ignored (in white) in the sum for the B-statistic computation with $n=10$.
	\emph{(Left)} $b=2$, $n_1=n_2=5$. 
	\emph{(Centre)} $b=2$, $n_1=3$, $n_2=5$. 
	\emph{(Right)} $b=3$, $n_1=4$, $n_2=4$, $n_3=2$. 
	}
\end{figure}

\paragraph{X-statistic.}
The cross X-statistic, introduced by \citealp{kim2023dimension} (see also \citealp{shekhar2022two,shekhar2022ind}), considers the entries
$\D \coloneqq \{(i,j): i = 1, \dots, n_1 \text{ for } j = n_1+1, \dots, n\}$ for some $n_1\in\{1,\dots,n-1\}$, giving
\begin{equation}
X = \frac{1}{n_1(n-n_1)}\sum_{i=1}^{n_1}\sum_{j=n_1+1}^n h(X_{i}, X_{j})
\end{equation}
computable in time complexity $\bigO{n_1(n-n_1)}$.
The main point of using this statistic is that the terms appearing in the first input of the core $h$ and in its second input, are disjoint.
Leveraging this fact, by scaling the statistic appropriately by some standard deviation (\ie studentisation), asymptotic normality of the statistic can always be guaranteed \citep{kim2023dimension}.
A typical choice for $n_1$ is simply to set it equal to $\lfloor n/2 \rfloor$, in which case the time complexity is still quadratic ($(n/2)^2$ rather than $n^2$) but this statistic can benefit from asymptotic normality.

\begin{figure}[H]
	\centerline{
	\includegraphics[height=5cm]{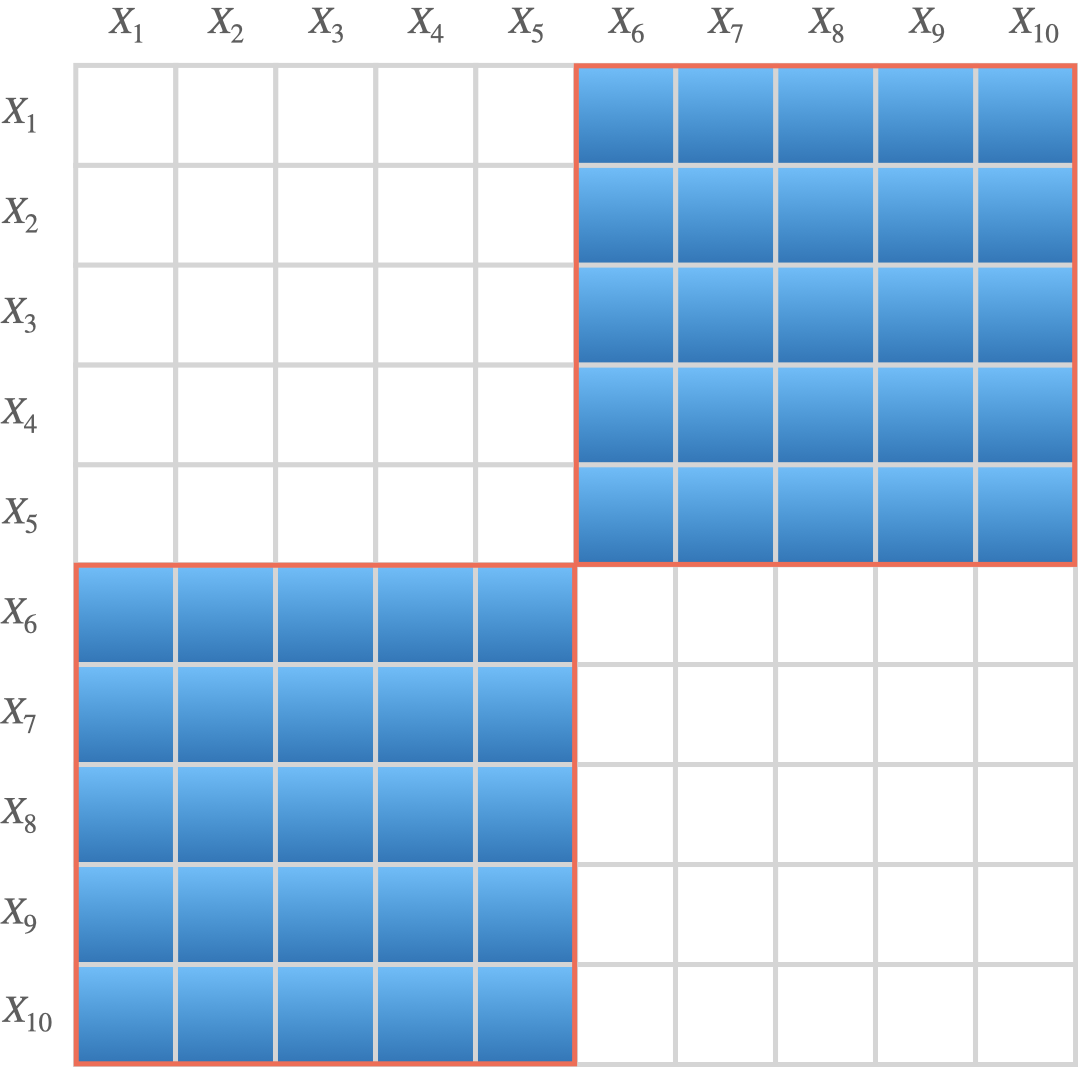}
	\hspace{0.5cm}
	\includegraphics[height=5cm]{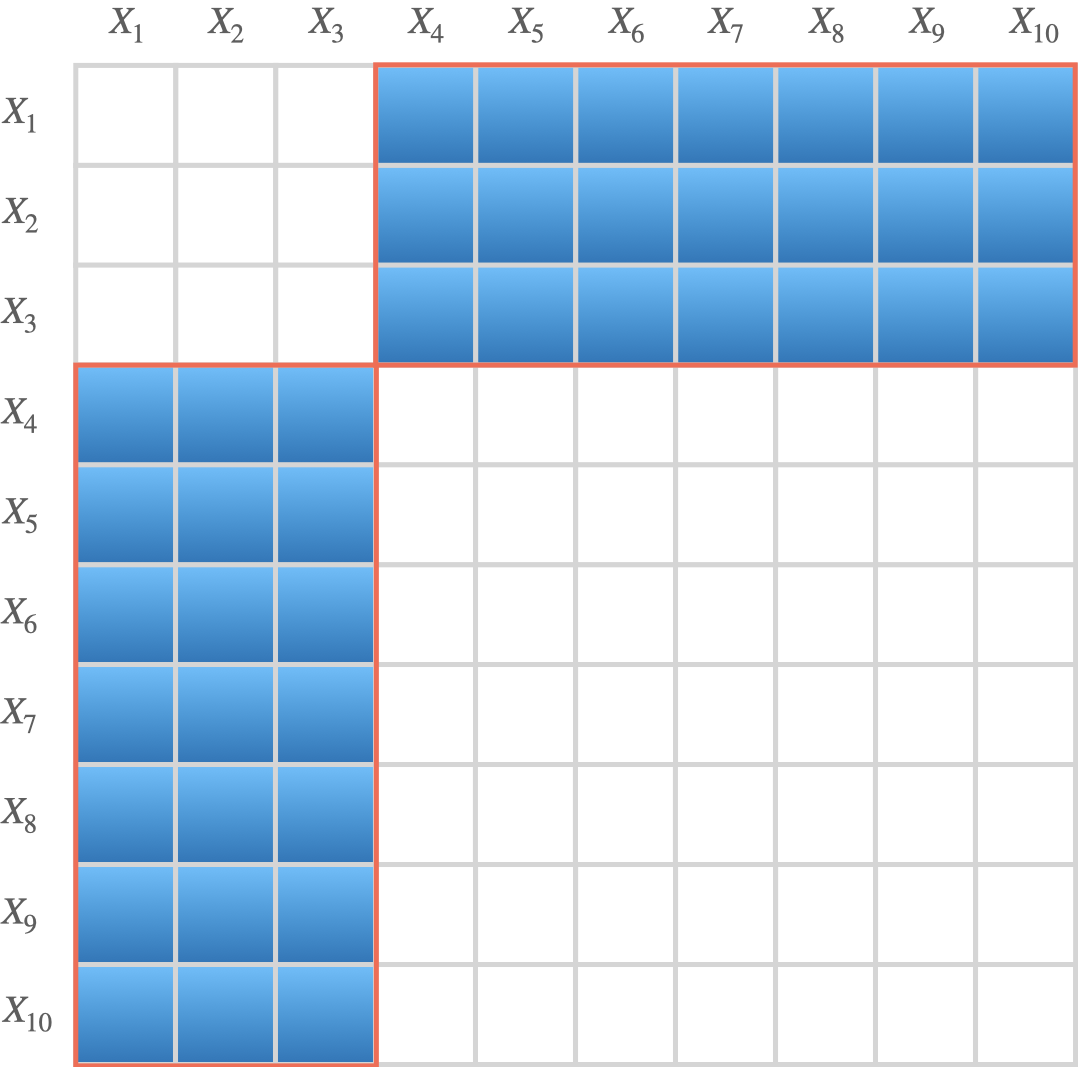}
	\hspace{0.5cm}
	\includegraphics[height=5cm]{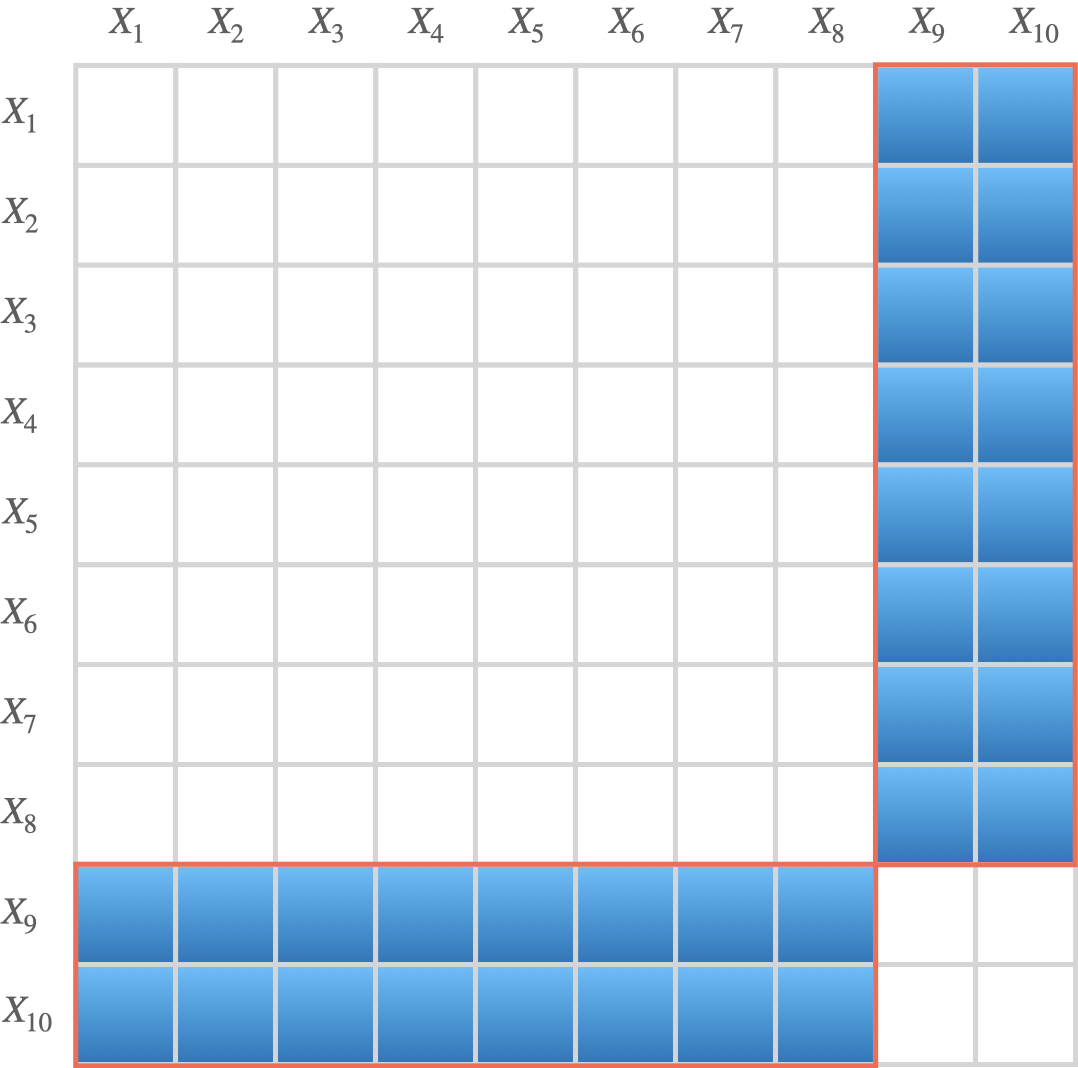}
	}
	\caption{X-statistic. Visualisation of the core kernel matrix entries $h(X_i,X_j)$ considered (in  blue) and ignored (in white) in the sum for the X-statistic computation with $n=10$.
	\emph{(Left)} $n_1=5$. 
	\emph{(Centre)} $n_1=3$. 
	\emph{(Right)} $n_1=8$. 
	}
\end{figure}

\paragraph{R-statistic.} 
Let $\Dcal_r$ be a random subsample of $\{(i,j): 1\leq i< j \leq n\}$, either with or without replacement, of some prespecified size $|\Dcal_r|$. 
Then, the random R-statistic \citep{lee1990ustatistic} is defined as
\begin{equation}
R=\frac{1}{|\Dcal_r|}\sum_{(i,j) \in \Dcal_r} h(X_i, X_j)
\end{equation}
with time complexity $\bigO{|\Dcal_r|}$ chosen by the user.
Given some fixed data $X_1,\dots,X_n$, computing the R-statistic twice results in different values due to the additional source of randomness introduced in the statistic computation.
This statistic has the benefit that in expectation it considers all non-diagonal entries of the core kernel matrix while being computationally faster to be evaluated.

\begin{figure}[H]
	\centerline{
	\includegraphics[height=5cm]{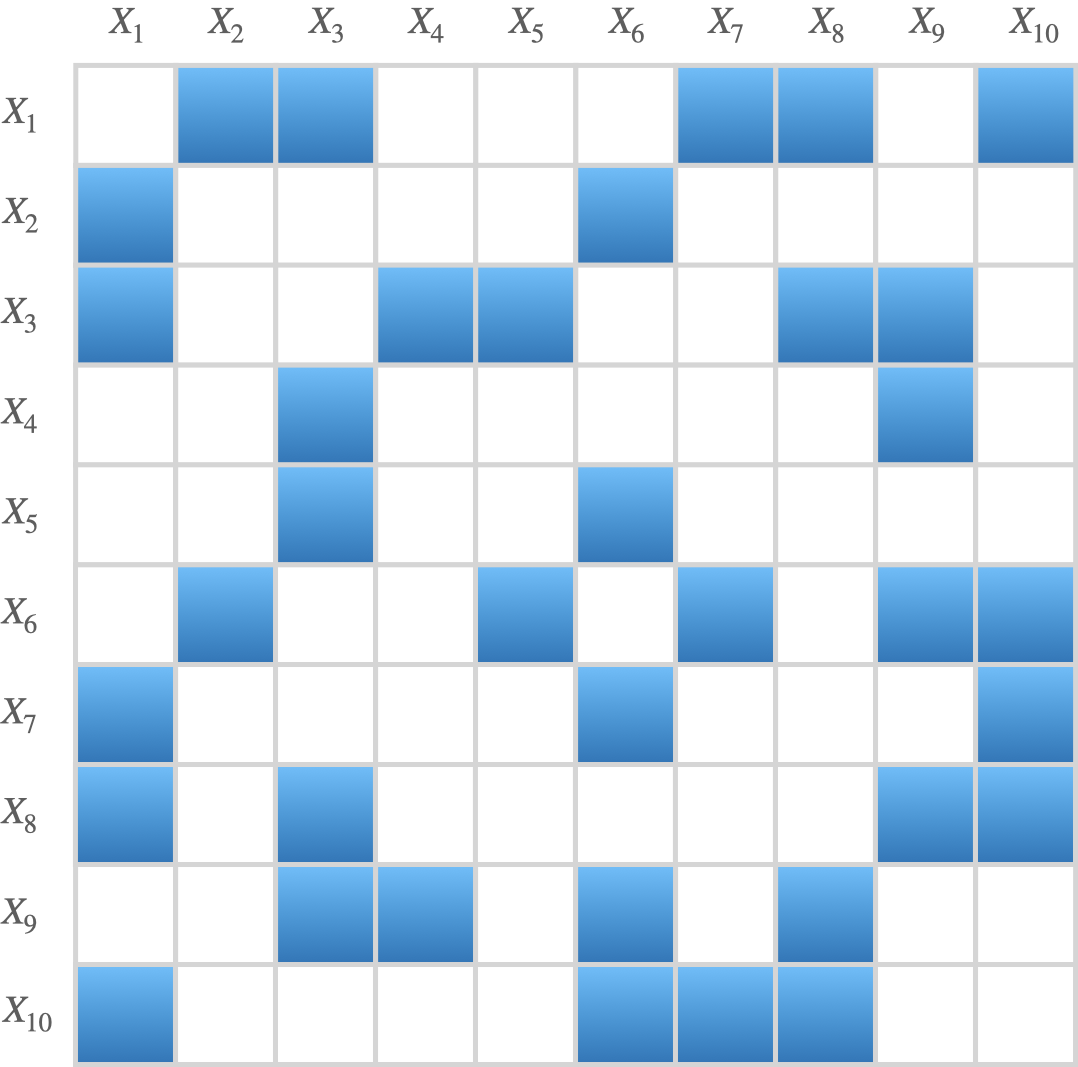}
	\hspace{0.5cm}
	\includegraphics[height=5cm]{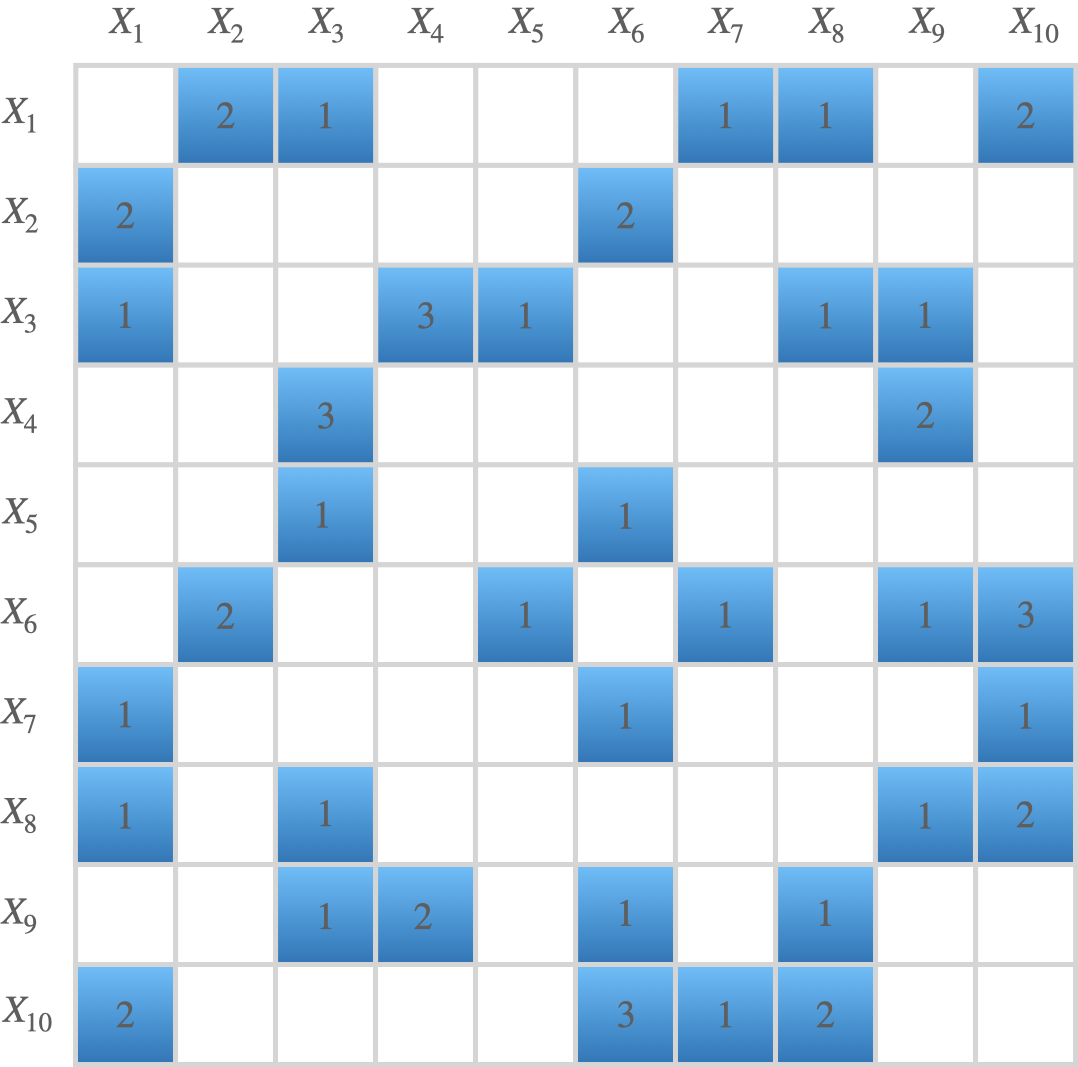}
	}
	\caption{R-statistic. Random visualisation of the core kernel matrix entries $h(X_i,X_j)$ sampled (in  blue) and not sampled (in white) in the sum for the R-statistic computation with $n=10$.
	\emph{(Left)} Without replacement, $|\Dcal_r| = 18$. 
	\emph{(Right)} With replacement---the numbers represent how many times each entry has been sampled in the upper triangular matrix, $|\Dcal_r| = 27$. 
	}
\end{figure}

\mysection{Kernel pooling: adaptive kernel discrepancies estimators}
\label{sec:adaptive_estimators}

\paragraph{Adaptivity via kernel pooling.} 
As aforementioned, the kernel choices for the MMD, HSIC and KSD estimators crucially affect their utility.
To overcome this, we rely on kernel pooling, which consists in combining multiple statistics with different kernels, to construct estimators which are adaptive to the kernel selection.
We first explain how the statistics can be normalised to be compared against each other, we then present three kernel pooling methods for combining them, and finally we propose a parameter-free method for constructing a collection of kernels. 
We note that kernel pooling can be used either with or without normalisation.
In practice, we recommend using fuse kernel pooling (\Cref{eq:fuse6}) with normalisation (\Cref{eq:normalised_statistic}), which is studied in depth in \citet{biggs2023mmd}.

\paragraph{Normalisation.} 
Consider a finite collection of kernels\footnote{For HSIC, this collection consists of product kernels $k=\hsickx\times\hsicky$.} $\Kcal$ and their associated statistics\footnote{For simplicity, we consider one-sample second-order statistics of this form, but the method holds more generally for any statistic (the normalisation needs to be adapted accordingly).}
$$
S_k 
= \frac{1}{|\Dcal_k|}\sum_{(i,j) \in \Dcal_k} h_k(X_i, X_j).
$$
In order to compare the statistics $S_1,\dots,S_{|\Kcal|}$, we need to ensure they are indeed comparable. 
For example, scaling the kernel by a constant trivially scales the MMD and HSIC by that factor.
Moreover, some statistics might have much higher variance than others.
These two facts illustrate that simply having a larger statistic might not necessarily be significant.
To account for this, we propose to normalise the estimators by some standard deviation term, that is, instead of considering $S_k$, we compute
\begin{equation}
\label{eq:normalised_statistic}
S_k / \sigma_k 
\qquad
\textrm{ where }
\qquad
\sigma_k^2 
\coloneqq
\frac{4}{\big|\Dcal_k^1\big|}\sum_{i \in \Dcal_k^{1}} 
\left(
\frac{1}{\big|\Dcal_k^{2,i}\big|}\sum_{j \in \Dcal_k^{2,i}}
h_k(X_i, X_j)
\right)^2
-
\
\left(
\frac{2}{\big|\Dcal_k\big|}\sum_{(i,j) \in \Dcal_k}
h_k(X_i, X_j)
\right)^2
\end{equation}
where $\Dcal_k^1\coloneqq \{i: (i,j)\in\Dcal_k\textrm{ for some } j\}$ and $\Dcal_k^{2,i} \coloneqq \{j:(i,j)\in\Dcal_k\}$. 
This biased standard deviation estimator of the statistic under the alternative hypothesis in this incomplete form has been adapted from the complete variant proposed by \citet{sutherland2016generative} and \citet[Equation 5]{liu2020learning}, see \citet{sutherland2022unbiased} for unbiased standard deviation estimators.

While this appears to be similar to studentisation, we emphasise that the aim is different: we are not interested in obtaining asymptotic normality but in being able to compare all the normalised statistics $S_1/\sigma_1,\dots,S_{|\Kcal|}/\sigma_{|\Kcal|}$ in a meaningful way.
We note that, for studentisation, there is no real consensus in the literature on which form the estimated standard deviation should take.
Here, we propose to use a simple one which aligns well with our study of different types of statistics in \Cref{sec:fixed_estimators}.
The unnormalised case simply corresponds to using $\sigma_k=1$.

We now present three methods for combining the (normalised) statistics, namely, mean, maximum and fuse kernel pooling.
These run in time complexity $\bigO{\sum_{k\in\Kcal}|\Dcal_k|}$ which is $\bigO{|\Kcal||\Dcal|}$ if the same design $\Dcal$ is used across all statistics.

\paragraph{Mean kernel pooling.} 
One possibility is to take the mean (or sometimes the sum) of the normalised statistics, giving
\begin{equation}
\underset{k\in\Kcal}{\mathrm{mean}}~S_k / \sigma_k
~=~
\frac{1}{|\Kcal|}\sum_{k\in\Kcal}\ S_k / \sigma_k .
\end{equation}
All normalised estimators are added up together, the intuition being that, as long as one statistic is `large', then this will be captured in the sum.

Another common method is to take the mean (or the sum) of kernels, and then to simply compute one statistic with this mean kernel.
For the case of MMD and HSIC, due to the linearity with respect to the kernel, we note that this is equivalent to taking the mean (or the sum) of the MMDs/HSICs without normalisation, that is
\begin{equation}
\label{eq:mean_kernel_mmd}
S_{\overline{k}}
=
\frac{1}{|\Kcal|}\sum_{k\in\Kcal}\ S_k
\qquad
\textrm{ where }
\qquad
{\overline{k}}
\coloneqq
\frac{1}{|\Kcal|}\sum_{k\in\Kcal} k.
\end{equation}

\paragraph{Maximum kernel pooling.} 
In order to capture the discrepancy, another possibility is simply to take the largest of the normalised statistics, that is, to compute
\begin{equation}
\max_{k\in\Kcal}\ S_k / \sigma_k. 
\end{equation}
Intuitively, if the maximum is `large', the normalised statistic is `large' for some kernel, meaning that the discrepancy can be detected for that kernel.
If the maximum is `small', we deduce that all statistics are `small' and that there is no discrepancy detected by any of the kernels.

In this method, only one value is retained, which differs from the previous method in which all values are combined.
The fact that many values are simply ignored, and that slightly modifying them might not change the maximum, can often not be desirable, both from a theoretical and a practical point of view.
The unnormalized version of maximum kernel pooling is very closely related to the methods of \citet{fukumizu2009kernel} and \citet{carcamo2022uniform}.
We next present a relaxed maximum which overcomes these issues. 

\paragraph{Fuse kernel pooling.} 
We can use a relaxed maximum of the normalised statistics which takes the form of a logsumexp expression
\begin{equation}
\label{eq:fuse6}
\underset{k\in\Kcal}{\mathrm{fuse}}~S_k / \sigma_k
~=~
\frac{1}{\nu} \log \left( \frac{1}{|\Kcal|}\sum_{k\in\Kcal} \exp \left(\nu S_k / \sigma_k \right)  \right)
\end{equation}
with fusing parameter $\nu>0$.
Noting that $\exp(\nu M)/|\Kcal| \leq \frac{1}{|\Kcal|}\sum_{k\in\Kcal} \exp(\nu S_k/\sigma_k) \leq \exp(\nu M)$ where $M\coloneqq\max_{k\in \Kcal} S_k/\sigma_k$, we deduce that
\begin{equation}
\label{eq:fuse_max}
\max_{k\in\Kcal} S_k /\sigma_k - \frac{\log(|\Kcal|)}{\nu}
~~\leq~~
\underset{k\in\Kcal}{\mathrm{fuse}}~S_k / \sigma_k
~~\leq~~
\max_{k\in\Kcal} S_k / \sigma_k 
\end{equation}
Hence, as $\nu$ tends to infinity, the estimator converges to the maximum of the $S_1/\sigma_1,\dots,S_{|\Kcal|}/\sigma_{|\Kcal|}$ values.
Building upon the theory of \citet{biggs2023mmd}, a typical choice of $\nu$ is to set it equal to ${\max_{k\in\Kcal}|\Dcal_k|/N}$ which increases with the sample size $N$ for estimators computable in time longer than linear.
For complete quadratic-time statistics, this gives $\nu = N$.
The same intuition as for the true maximum holds, but having an estimator which changes with each normalised statistic can be beneficial for downstream tasks.
As illustrated in \citet[Appendix B]{biggs2023mmd}, fuse kernel pooling also allows for (uncountable) distributions on the space of kernels instead of simply working with a finite collection of kernels (\ie, uniform distribution on a discrete set of kernels).
In practice, we recommend using the fuse variant of kernel pooling, which is analyzed in details in \citet{biggs2023mmd}.

\paragraph{Kernel collection.}
Consider a radial kernel
$
k_\lambda(x,y) = \Psi\pp{{\|x-y\|_r}/{\lambda}}
$
for some $r\geq 1$, $\lambda>0$ and $\Psi\colon \R\to\R$ normalised (either such that it integrates to $1$, or such that $\psi(0)=1$).
Given some data $X_1,\dots,X_n$, consider the set of inter-sample distances
\begin{equation}
D = 
\big\{
\|x-x'\|_r
\colon
x,x'\in\{X_1,\dots,X_n\},\, x\neq x'
\big\}\setminus\{0\}.
\end{equation}
A naive way of choosing the kernel bandwidth is simply to set it equal to the median of $D$ \citep{gretton2012kernel}, while simple, this method fails to capture the discrepancy accurately in most cases and is not adaptive.
However, the set $D$ of distances remains very relevant as the kernel is evaluated at these values scaled by the inverse bandwidth.
Hence, to construct a collection of bandwidths for $k$ from $D$ it makes sense to consider a discretisation of the interval between the minimum and maximum of $D$. 
In practice, to avoid numerical issues, we actually use the 5\% and 95\% quantiles of $D$ instead, and discretise the interval between them linearly using 10 points (\citealp[Section 6]{biggs2023mmd}). 
As noted in \citet[Section 5.7]{schrab2021mmd}, using only 10 bandwidths is sufficient to fully capture all the information, no advantage is observed for using more points in the discretisation.
For the kernel $k$, this gives a bandwidth collection
\begin{equation}
\Lambda(k) = \big\{q_{5\%} + i (q_{95\%} -q_{5\%}) / 9 : i = 0,\dots, 9\big\}.
\end{equation}
To construct the collection of kernels, we can then consider multiple kernel types and use for each the 10 bandwidths constructed above.
In practice, as illustrated in \citet[Section 5.7]{schrab2021mmd}, we recommend combining Gaussian and Laplace kernels, with no advantage observed for including more types of kernels.
The parameter-free kernel collection is then
\begin{equation}
\label{eq:collection_fuse}
\Kcal = \big\{
k_\lambda \colon
k\in \{\textrm{Gaussian, Laplace}\},
\lambda\in \Lambda(k)
\big\}
\end{equation}
consisting of 20 kernels, which can then be used when computing an adaptive estimator through mean, maximum or fuse kernel pooling.
In practice, when using these kernel metrics in general settings, we recommend using fuse pooling studied in details in \citet{biggs2023mmd}.
When using them for hypothesis testing, another powerful adaptive method is \emph{aggregation} \citep{schrab2021mmd,schrab2022ksd,albert2019adaptive}.
See \citet{schrab2025unified} for a unified view of hypothesis testing optimality results using these kernel discprepancies.

\fancyhf{}
\fancyfoot[RO]{\thepage}
\lhead{}
\chead{A Practical Introduction to Kernel Discrepancies: MMD, HSIC \& KSD}
\cfoot{\thepage}
\rfoot{}
\bibliography{biblio}
\section*{Acknowledgements}
I, Antonin Schrab, acknowledge support from the U.K.\ Research and Innovation under grant number EP/S021566/1.

\end{document}